# The Divide-and-Conquer Subgoal-Ordering Algorithm for Speeding up Logic Inference

**Oleg Ledeniov**                                        OLLEG@CS.TECHNION.AC.IL
**Shaul Markovitch**                                     SHAULM@CS.TECHNION.AC.IL
*Computer Science Department*
*Technion – Israel Institute of Technology*
*Haifa 32000, Israel*

## Abstract

It is common to view programs as a combination of logic and control: the logic part defines *what* the program must do, the control part – *how* to do it. The Logic Programming paradigm was developed with the intention of separating the logic from the control. Recently, extensive research has been conducted on automatic generation of control for logic programs. Only a few of these works considered the issue of automatic generation of control for improving the efficiency of logic programs. In this paper we present a novel algorithm for automatic finding of lowest-cost subgoal orderings. The algorithm works using the divide-and-conquer strategy. The given set of subgoals is partitioned into smaller sets, based on co-occurrence of free variables. The subsets are ordered recursively and merged, yielding a provably optimal order. We experimentally demonstrate the utility of the algorithm by testing it in several domains, and discuss the possibilities of its cooperation with other existing methods.

## 1. Introduction

It is common to view programs as a combination of logic and control (Kowalski, 1979). The logic part defines *what* the program must do, the control part – *how* to do it. Traditional programming languages require that the programmers supply both components. The Logic Programming paradigm was developed with the intention of separating the logic from the control (Lloyd, 1987). The goal of the paradigm is that the programmer specifies the logic without bothering about the control, which should be supplied by the interpreter.

Initially, most practical logic programming languages, such as Prolog (Clocksin & Mellish, 1987; Sterling & Shapiro, 1994), did not include the means for automatic generation of control. As a result, a Prolog programmer had to implicitly define the control by the order of clauses and of subgoals within the clauses. Recently, extensive research has been conducted on automatic generation of control for logic programs. A major part of this research is concerned with control that affects correctness and termination of logic programs (De Schreye & Decorte, 1994; Somogyi, Henderson, & Conway, 1996b; Cortesi, Le Charlier, & Rossi, 1997). Only a few of these works consider the issue of automatic generation of control for improving the *efficiency* of logic programs. Finding a good ordering that leads to efficient execution requires a deep understanding of the logic inference mechanism. Hence, in many cases, only expert programmers are able to generate efficient programs. The problem intensifies with the recent development of the field of inductive logic programming (Muggleton

& De Raedt, 1994). There, logic programs are automatically induced by learning. Such learning algorithms are commonly built with the aim of speeding up the induction process without considering the efficiency of resulting programs.

The goal of the research described in this paper is to design algorithms that automatically find efficient orderings of subgoal sequences. Several researchers have explored the problem of automatic reordering of subgoals in logic programs (Warren, 1981; Naish, 1985b; Smith & Genesereth, 1985; Natarajan, 1987; Markovitch & Scott, 1989). The general subgoal ordering problem is known to be NP-hard (Ullman, 1982; Ullman & Vardi, 1988). Smith and Genesereth (1985) and Markovitch and Scott (1989) present search algorithms for finding optimal orderings. These algorithms are general and carry exponential costs for non-trivial sets of subgoals. Natarajan (1987) describes an efficient algorithm for the special case where subgoals in the set do not share free variables.

In this paper we present a novel algorithm for subgoal ordering. We call two subgoals that share a free variable *dependent*. Unlike Natarajan's approach, which can only handle subgoal sets that are completely independent, our algorithm can deal with any subgoal set, while making maximal use of the existing dependencies for acceleration of the ordering process. In the worst case the algorithm – like that of Smith and Genesereth – is exponential. Still, in most practical cases, our algorithm exploits subgoal dependencies and finds optimal orderings in polynomial time.

We start with an analysis of the ordering problem and demonstrate its importance through examples. We then show how to compute the cost of a given ordering based on the cost and the number of solutions of the individual subgoals. We describe the algorithm of Natarajan and the algorithm of Smith and Genesereth and show how the two can be combined into an algorithm that is more efficient and general than each of the two. We show drawbacks of the combined algorithm and introduce the new algorithm, which avoids these drawbacks. We call it the *Divide-and-Conquer algorithm (*DAC *algorithm)*. We prove the correctness of the algorithm, discuss its complexity and compare it to the combined algorithm. The DAC algorithm assumes knowledge of the cost and the number of solutions of the subgoals. This knowledge can be obtained by machine learning techniques such as those employed by Markovitch and Scott (1989). Finally, we test the utility of our algorithm by running a set of experiments on artificial and real domains.

The DAC algorithm for subgoal ordering can be combined with many existing methods in logic programming, such as program transformation, compilation, termination control, correctness verification, and others. We discuss the possibilities of such combinations in the concluding section.

Section 2 states the ordering problem. Section 3 describes existing ordering algorithms and their combination. Section 4 presents the new algorithm. Section 5 discusses the acquisition of the control knowledge. Section 6 contains experimental results. Section 7 contains a discussion of practical issues, comparison with other works and conclusions.

## 2. Background: Automatic Ordering of Subgoals

We start by describing the conventions and assumptions accepted in this paper. Then we demonstrate the importance of subgoal ordering and discuss its validity. Finally, we present a classification of ordering methods and discuss related work.

## 2.1 Conventions and Assumptions

All constant, function and predicate symbols in programs begin with lower case letters, while capital letters are reserved for variables. Braces are used to denote unordered sets (e.g., $\{a, b, c\}$), and angle brackets are used for ordered sequences (e.g., $\langle a, b, c \rangle$). Parallel lines ($\|$) denote concatenations of ordered sequences of subgoals. When speaking about abstract subgoals (and not named predicates of concrete programs), we denote separate subgoals by capital letters ($A, B \ldots$), ordered sequences of subgoals by capitalized vectors ($\vec{B}, \vec{O_S} \ldots$), and sets of subgoals by calligraphic capitals ($\mathcal{B}, \mathcal{S} \ldots$). $\pi(\mathcal{S})$ denotes the set of all permutations of $\mathcal{S}$.

We assume that the programs we work with are written in *pure Prolog*, i.e., without cut operators, meta-logical or extra-logical predicates. Alternatively, we can assume that only pure Prolog sub-sequences of subgoals are subject to ordering. For example, given a rule of the form

$$A \leftarrow B_1, B_2, B_3, !, B_4, B_5, B_6.$$

only its final part $\{B_4, B_5, B_6\}$ can be ordered (without affecting the solution set).

In this work we focus upon the task of finding *all* the solutions to a set of subgoals.

## 2.2 Ordering of Subgoals in Logic Programs

A logic program is a set of clauses:

$$A \leftarrow B_1, B_2, \ldots, B_n. \qquad (n \geq 0)$$

where $A, B_1, \ldots, B_n$ are literals (predicates with arguments). To use such a clause for proving a goal that matches $A$, we must prove that *all* $B$-s hold simultaneously, under consistent bindings of the free variables. A *solution* is such a set of variable bindings. The *solution set* of a goal is the bag of all its solutions created by its program.

A *computation rule* defines which subgoal will be proved next. In Prolog, the computation rule always selects the leftmost subgoal in a goal. If a subgoal fails, *backtracking* is performed – the proof of the previous subgoal is re-entered to generate another solution. For a detailed definition of the logic inference process, see Lloyd (1987).

**Theorem 1** *The solution set of a set of subgoals does not depend on the order of their execution.*

**Proof:** When we are looking for all solutions, the solution set does not depend on the computation rule chosen (Theorems 9.2 and 10.3 in Lloyd, 1987). Since a transposition of subgoals in an ordered sequence can be regarded as a change of the computation rule (the subgoals are selected in different order), such transposition does not change the solution set. □

This theorem implies that we may reorder subgoals during the proof derivation. Yet the *efficiency* of the derivation strongly depends on the chosen order of subgoals. The following example illustrates how two different orders can lead to a large difference in execution efficiency.

```
        parent(abraham,isaac).        male(abraham).
        parent(sarah,isaac).          male(isaac).
        parent(abraham,ishmael).      male(ishmael).
        parent(isaac,esav).           male(jakov).
        parent(isaac,jakov).          male(esav).
        ... More parent clauses ...   ... More male clauses ...

 brother(X,Y) ← male(X), parent(W,X), parent(W,Y), X=/=Y.
 father(X,Y) ← male(X), parent(X,Y).
 uncle(X,Y) ← parent(Z,Y), brother(X,Z).
 ... More rules of relations ...
```

Figure 1: A small fragment of a Biblical database describing family relationships.

## Example 1

Consider a Biblical family database such as the one listed in Figure 1 (a similar database appears in the book by Sterling & Shapiro, 1994). The body of the rule defining the uncle-nephew (or uncle-niece) relation can be ordered in two ways:

1. `uncle(X,Y) ← brother(X,Z), parent(Z,Y).`

2. `uncle(X,Y) ← parent(Z,Y), brother(X,Z).`

To prove the goal `uncle(ishmael,Y)` using the first version of the rule, the interpreter will first look for Ishmael's siblings (and find Isaac) and then for the siblings' children (Esav and Jacov). The left part of Figure 2 shows the associated proof tree with a total of 10 nodes. If we use the second version of the rule, the interpreter will create all the parent-child pairs available in the database, and will test for each parent whether he (or she) is Ishmael's sibling. The right part of Figure 2 shows the associated proof tree with a total of $4(N-2) + 6 \cdot 2 + 2 = 4N + 6$ nodes, where N is the number of parent-child pairs in the database. The tree contains two success branches and $N-2$ failure branches; in the figure we show one example of each. While the two versions of the rule yield identical solution sets, the first version leads to a much smaller tree and to a faster execution.

Note that this result is true only for the given mode (`bound,free`) of the head literal; for the mode (`free,bound`), as in `uncle(X,jacov)`, the outcome is the contrary: the second version of the rule yields a smaller tree.

## 2.3 Categories of Subgoal Ordering Methods

Assume that the current conjunctive goal (the current resolvent) is $\{A_1, A_2\}$. Assume that we use the rule "$A_1 \leftarrow A_{11}, A_{12}$." to reduce $A_1$. According to Theorem 1, the produced resolvent, $\{A_{11}, A_{12}, A_2\}$, can be executed in any order. We call ordering methods that allow any permutation of the resolvent *interleaving ordering methods*, since they permit interleaving of subgoals from different rule bodies. When ordering is performed only on rule bodies *before* using them for reduction, the method is *non-interleaving*. In the above example, interleaving methods will consider all 6 permutations of the resolvent, while non-interleaving methods will consider only two orderings: $\langle A_{11}, A_{12}, A_2 \rangle$ and $\langle A_{12}, A_{11}, A_2 \rangle$. Interleaving ordering methods deal with significantly more possible orderings than non-interleaving methods. That means that they can find more efficient orderings. On the other hand, the space of possible orderings may become prohibitively large, requiring too many computational resources.

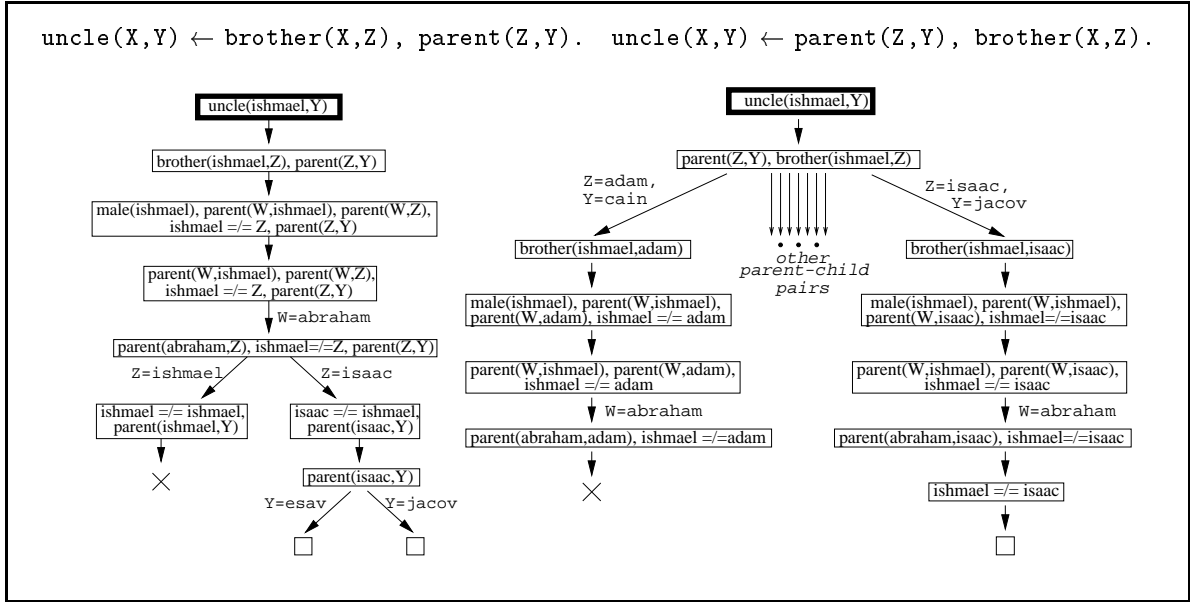

Figure 2: Two proof trees obtained with different orderings of a single rule in Example 1.

Subgoal ordering can take place at various stages of the proof process. We divide all subgoal ordering methods into static, semi-dynamic and dynamic.

- **Static ordering:** The rule bodies are ordered before the execution starts. No ordering takes place during the execution.

- **Semi-dynamic ordering:** Whenever a rule is selected for reduction, its body is ordered. The order of its subgoals does not change after the reduction takes place.

- **Dynamic ordering:** The ordering decision is made at each inference step.

Static methods add no overhead to the execution time. However, the optimal ordering of a rule often depends on a particular binding of a variable, which can be known only at run-time. For instance, in Example 1 we saw that the first ordering of the rule is better for proving the goal uncle(ishmael,Y). And yet, for the goal uncle(X,jacov), it is the second ordering that yields more efficient execution. To handle such cases statically, we must compute the optimal ordering for each possible binding.

Obviously, static ordering can only be non-interleaving. The dynamic method is more flexible, since it can use more updated knowledge about variable bindings, but it also carries the largest runtime overhead, since it is invoked several times for each use of a rule body. The semi-dynamic method is a compromise between the two: it is more powerful than the static method, because it can dynamically propose different orderings for different instances of the same rule; it also carries less overhead than the dynamic method, because it is invoked only once for each use of a rule body.

The total time of proving a goal is the sum of the ordering time and the inference time. Interleaving and dynamic methods have the best potential for reducing the inference time, but may significantly augment the ordering time. Static methods do not devote time to ordering (it is done off-line), but have a limited potential for reducing the inference time.

The algorithms described in this paper can be used for all categories of ordering methods, although in the experiments described in Section 6 we have only implemented semi-dynamic, non-interleaving ordering methods: on each reduction, the rule body is ordered and added to the left end of the resolvent, and then the leftmost literal of the resolvent is selected for the next reduction step.

## 2.4 Related Work

The problem of computational inefficiency of logic inference was the subject of extensive research. The most obvious aspect of this inefficiency is the possible non-termination of a proof. Several researchers developed compile-time and run-time techniques to detect and avoid infinite computations (De Schreye & Decorte, 1994). A certain success was achieved in providing more advanced control through employment of co-routining for inter-predicate synchronization purposes (Clark & McCabe, 1979; Porto, 1984; Naish, 1984). Also, infinite computations can be avoided by pruning infinite branches that do not contain solutions (Vasak & Potter, 1985; Smith, Genesereth, & Ginsberg, 1986; Bol, Apt, & Klop, 1991). In the NAIL! system (Morris, 1988) subgoals are automatically reordered to avoid nontermination.

Still, even when the proof is finite, it is desirable to make it more efficient. Several researchers studied the problem of clause ordering (Smith, 1989; Cohen, 1990; Etzioni, 1991; Laird, 1992; Mooney & Zelle, 1993; Greiner & Orponen, 1996). If we are looking for *all* the solutions of a goal, then the efficiency does not depend on the clause order (assuming no `cuts`). Indeed, if some predicate has $m$ clauses, and for some argument bindings these clauses produce all their solutions in times $t_1, t_2 \ldots t_m$, then all solutions of the predicate under these bindings are obtained in time $t_1 + t_2 + \ldots + t_m$, regardless of the order in which the clauses are applied. Different clause orderings correspond to different orders in which branches are selected in a proof tree; if we traverse the entire tree, then the number of traversal steps does not depend on the order of branch selection, though the order of solutions found does depend on it.

Subgoal ordering, as was demonstrated in Example 1, can significantly affect the efficiency of proving a goal. There are two major approaches to subgoal ordering. The first approach uses various heuristics to order subgoals, for example:

- Choose a subgoal whose predicate has the smallest number of matching clauses (Minker, 1978).

- Prefer a subgoal with more constants (Minker, 1978).

- Choose a subgoal with the largest *size*, where the size is defined as the number of occurrences of predicate symbols, function symbols, and variables (Nie & Plaisted, 1990).

- Choose a subgoal with the largest *mass*, where the mass of a subgoal depends on the frequency of its arguments and sub-arguments in the entire goal (Nie & Plaisted, 1990).

- Choose a subgoal with the least number of solutions (Warren, 1981; Nie & Plaisted, 1990).

- Apply "tests" before "generators" (Naish, 1985a).

- Prefer calls that fail quickly (Naish, 1985b).

The heuristic methods usually execute quickly, but may yield suboptimal orderings.

The second approach, which is adopted in this paper, aims at finding *optimal* orderings (Smith & Genesereth, 1985; Natarajan, 1987; Markovitch & Scott, 1989). Natarajan proposed an efficient way to order a special sort of subgoal set (where all subgoals are independent), while Smith and Genesereth proposed a general, but inefficient algorithm. In the following section we build a unifying framework for dealing with subgoal ordering and describe variations on Natarajan's and Smith and Genesereth's algorithms. We also show how the two can be combined for increased efficiency.

## 3. Algorithms for Subgoal Ordering in Logic Programs

The goal of the work presented here is to order subgoals for speeding up logic programs. This section starts with an analysis of the cost of executing a sequence of subgoals. The resulting formula is the basis for the subsequent ordering algorithms. Then we discuss dependence of subgoals and present existing ordering algorithms for independent and dependent sets of subgoals. Finally, we combine these algorithms into a more general and efficient one.

### 3.1 The Cost of Executing a Sequence of Subgoals

In this subsection we analyze the cost of executing a sequence of subgoals. The analysis builds mainly on the work of Smith and Genesereth (1985).

Let $\mathcal{S} = \{A_1, A_2, \ldots A_k\}$ be a set of subgoals and $b$ be a binding. We denote $Sols(\mathcal{S})$ to be the solution set of $\mathcal{S}$, and define $Sols(\emptyset) = \{\emptyset\}$. We denote $A_i|_b$ to be $A_i$ whose variables are bound according to $b$ ($A_i|_\emptyset = A_i$). Finally, we denote $Cost(A_i|_b)$ to be the amount of resources needed for proving $A_i|_b$. $Cost(A_i|_b)$ should reflect the time complexity of proving $A_i$ under binding $b$. For example, the number of unification steps is a natural measure of complexity for logic programs (Itai & Makowsky, 1987).

To obtain the cost of finding all the solutions of an ordered sequence of subgoals

$$\vec{S} = \langle A_1, A_2, A_3, \ldots A_n \rangle, \tag{1}$$

we note that the proof-tree of $A_1$ is traversed only once, the tree of $A_2$ is traversed once for each solution generated by $A_1$, the tree of $A_3$ – once for each solution of $\{A_1, A_2\}$, etc. Consequently, the total cost of proving Equation 1 is

$$
\begin{aligned}
Cost(\langle A_1, \ldots A_n \rangle) &= Cost(A_1) + \sum_{b \in Sols(\{A_1\})} Cost(A_2|_b) + \ldots + \sum_{b \in Sols(\{A_1, \ldots A_{n-1}\})} Cost(A_n|_b) = \\
&= \sum_{i=1}^{n} \sum_{b \in Sols(\{A_1, \ldots A_{i-1}\})} Cost(A_i|_b).
\end{aligned} \tag{2}
$$

To compute Equation 2 one must know the cost and the solution set for each subgoal under each binding. To reduce the amount of information needed, we derive an equivalent formula, which uses average cost and average number of solutions.

**Definition:** Let $\mathcal{B}$ be a set of subgoals, $A$ a subgoal. Define $c\overline{os}t(A)|_\mathcal{B}$ to be the average cost of $A$ over all solutions of $\mathcal{B}$ and $n\overline{so}ls(A)|_\mathcal{B}$ to be its average number of solutions over all solutions of $\mathcal{B}$:

$$
c\overline{os}t(A)|_\mathcal{B} = \begin{cases} Cost(A), & \mathcal{B} = \emptyset \\ \frac{\sum_{b \in Sols(\mathcal{B})} Cost(A|_b)}{|Sols(\mathcal{B})|}, & \mathcal{B} \neq \emptyset, \ Sols(\mathcal{B}) \neq \emptyset \\ undefined, & \mathcal{B} \neq \emptyset, \ Sols(\mathcal{B}) = \emptyset \end{cases}
$$

$$
n\overline{so}ls(A)|_\mathcal{B} = \begin{cases} |Sols(\{A\})|, & \mathcal{B} = \emptyset \\ \frac{\sum_{b \in Sols(\mathcal{B})} |Sols(\{A|_b\})|}{|Sols(\mathcal{B})|}, & \mathcal{B} \neq \emptyset, \ Sols(\mathcal{B}) \neq \emptyset \\ undefined, & \mathcal{B} \neq \emptyset, \ Sols(\mathcal{B}) = \emptyset \end{cases}
$$

From the first definition, it follows that:

$$
\sum_{b \in Sols(\{A_1, \ldots A_{i-1}\})} Cost(A_i|_b) = |Sols(\{A_1, \ldots A_{i-1}\})| \times c\overline{os}t(A_i)|_{\{A_1, \ldots A_{i-1}\}}. \tag{3}
$$

If we apply the second definition recursively, we obtain

$$
\begin{aligned}
|Sols(\{A_1, \ldots A_i\})| &= \sum_{b \in Sols(\{A_1, \ldots A_{i-1}\})} |Sols(\{A_i|_b\})| \\
&= |Sols(\{A_1, \ldots A_{i-1}\})| \times n\overline{so}ls(A_i)|_{\{A_1, \ldots A_{i-1}\}} \\
&= \ldots = \prod_{j=1}^{i} n\overline{so}ls(A_j)|_{\{A_1 \ldots A_{j-1}\}}. 
\end{aligned} \tag{4}
$$

Note that we defined $Sols(\emptyset) = \{\emptyset\}$; thus, these equations hold also for $i = 1$. Incorporation of Equations 3 and 4 into Equation 2 yields

$$
Cost(\langle A_1, A_2, \ldots A_n \rangle) = \sum_{i=1}^{n} \left[ \left( \prod_{j=1}^{i-1} n\overline{so}ls(A_j)|_{\{A_1 \ldots A_{j-1}\}} \right) \times c\overline{os}t(A_i)|_{\{A_1 \ldots A_{i-1}\}} \right]. \tag{5}
$$

For each subgoal $A_i$, its average cost is multiplied by the total number of solutions of all the preceding subgoals. We can define average cost and number of solutions for every continuous sub-sequence of Equation 1: $\forall k_1, k_2, 1 \leq k_1 \leq k_2 \leq n$,

$$c\overline{os}t(\langle A_{k_1}, \ldots A_{k_2} \rangle)|_{\{A_1, \ldots A_{k_1-1}\}} = \frac{c\overline{os}t(\langle A_1, \ldots A_{k_2} \rangle)|_{\emptyset} - c\overline{os}t(\langle A_1, \ldots A_{k_1-1} \rangle)|_{\emptyset}}{n\overline{sol}s(\langle A_1, \ldots A_{k_1-1} \rangle)|_{\emptyset}} \quad (6)$$

$$= \sum_{i=k_1}^{k_2} \left[ \left( \prod_{j=k_1}^{i-1} n\overline{sol}s(A_j)|_{\{A_1, \ldots A_{j-1}\}} \right) \times c\overline{os}t(A_i)|_{\{A_1, \ldots A_{i-1}\}} \right]$$

$$n\overline{sol}s(\langle A_{k_1}, \ldots A_{k_2} \rangle)|_{\{A_1, \ldots A_{k_1-1}\}} = \frac{n\overline{sol}s(\langle A_1, \ldots A_{k_2} \rangle)|_{\emptyset}}{n\overline{sol}s(\langle A_1, \ldots A_{k_1-1} \rangle)|_{\emptyset}} = \prod_{i=k_1}^{k_2} n\overline{sol}s(A_i)|_{\{A_1, \ldots A_{i-1}\}} \quad (7)$$

The values of $c\overline{os}t(A_i)$ and $n\overline{sol}s(A_i)$ depend on the position of $A_i$ in the ordered sequence. For example, assume that we want to find Abraham's sons, using the domain of Example 1. The unordered conjunctive goal is {male(Y),parent(abraham,Y)}. Let there be $N$ males in the database (two of them, Isaac and Ishmael, are Abraham's sons):

$$n\overline{sol}s(\texttt{male(Y)})|_{\emptyset} = N \qquad\qquad n\overline{sol}s(\texttt{parent(abraham,Y)})|_{\emptyset} = 2$$
$$n\overline{sol}s(\texttt{male(Y)})|_{\{\texttt{parent(abraham,Y)}\}} = 1 \quad n\overline{sol}s(\texttt{parent(abraham,Y)})|_{\{\texttt{male(Y)}\}} = 2/N$$

Note that $n\overline{sol}s(\langle \texttt{male(Y)},\texttt{parent(abraham,Y)} \rangle) = 2 = n\overline{sol}s(\langle \texttt{parent(abraham,Y)},\texttt{male(Y)} \rangle)$, exactly as Theorem 1 predicts.

Having defined the cost of a sequence of subgoals, we can now define the objective of our ordering algorithms:

**Definition:** Let $\mathcal{S}$ be a set of subgoals. Define $\pi(\mathcal{S})$ to be set of all permutations of $\mathcal{S}$. $\vec{O}_S \in \pi(\mathcal{S})$ is a *minimal ordering* of $\mathcal{S}$ (denoted $Min(\vec{O}_S, \mathcal{S})$), if its cost according to Equation 5 is minimal over all possible permutations of $\mathcal{S}$:

$$Min(\vec{O}_S, \mathcal{S}) \iff \forall O'_S \in \pi(\mathcal{S}) : Cost(\vec{O}_S) \leq Cost(O'_S).$$

The total execution time is the sum of the time which is spent on ordering, and the inference time spent by the interpreter on the ordered sequence. In this paper we focus upon developing algorithms for minimizing the inference time. Elsewhere (Ledeniov & Markovitch, 1998a, 1998b) we present algorithms that attempt to reduce the total execution time.

The values of cost and number of solutions can be obtained in various ways: by exact computation, by estimation and bounds, and by learning. Let us assume at the moment that there exists a mechanism that returns the average cost and number of solutions of a subgoal in time $\tau$. In Section 5 we show how this control knowledge can be obtained by inductive learning.

## 3.2 Ordering of Independent Sets of Subgoals

The general subgoal ordering problem is NP-hard (Ullman & Vardi, 1988). However, there is a special case where ordering can be performed efficiently: if all the subgoals in the given set are independent, i.e. do not share free variables. This section begins with the definition of subgoal dependence and related concepts. We then show an ordering algorithm for independent sets and prove its correctness.

### 3.2.1 DEPENDENCE OF SUBGOALS

**Definition:** Let $\mathcal{S}$ and $\mathcal{B}$ be sets of subgoals ($\mathcal{B}$ is called the *binding set* of $\mathcal{S}$). A pair of subgoals in $\mathcal{S}$ is *directly dependent* under $\mathcal{B}$, if they share a free variable not bound by a subgoal of $\mathcal{B}$.

A pair of subgoals is *indirectly dependent* with respect to $\mathcal{S}$ and $\mathcal{B}$ if there exists a third subgoal in $\mathcal{S}$ which is directly dependent on one of them under $\mathcal{B}$, and dependent (directly or indirectly) on the other one under $\mathcal{B}$. A pair of subgoals of $\mathcal{S}$ is *independent* under $\mathcal{B}$ if it is not dependent under $\mathcal{B}$ (either directly or indirectly). A subgoal is *independent* of $\mathcal{S}$ under $\mathcal{B}$ if it is independent of all members of $\mathcal{S}$ under $\mathcal{B}$.

Two subsets $\mathcal{S}_1 \subset \mathcal{S}$ and $\mathcal{S}_2 \subset \mathcal{S}$ are *mutually independent* under the binding set $\mathcal{B}$ if every pair of subgoals $(A_1, A_2)$, such that $A_1 \in \mathcal{S}_1$ and $A_2 \in \mathcal{S}_2$, is independent under $\mathcal{B}$.

The entire set $\mathcal{S}$ is called *independent* under the binding set $\mathcal{B}$ if all its subgoal pairs are independent under $\mathcal{B}$, and is called *dependent* otherwise. A dependent set of subgoals is called *indivisible* if all its subgoal pairs are dependent under $\mathcal{B}$, and *divisible* otherwise.

A *divisibility partition* of $\mathcal{S}$ under $\mathcal{B}$, $DPart(\mathcal{S}, \mathcal{B})$, is a partition of $\mathcal{S}$ into subsets that are mutually independent and indivisible under $\mathcal{B}$, except at most one subset which contains all the subgoals independent of $\mathcal{S}$ under $\mathcal{B}$. It is easy to show that $DPart(\mathcal{S}, \mathcal{B})$ is unique.

For example, let $\mathcal{S}_0 = \{a, b(X), c(Y), d(X,Y), e(Z), f(Z,V), h(W)\}$. With respect to $\mathcal{S}_0$ and an empty binding set, the pair $\{b(X), d(X,Y)\}$ is directly dependent, $\{b(X), c(Y)\}$ is indirectly dependent and $\{b(X), e(Z)\}$ is independent. If we represent a set of subgoals as a graph, where subgoals are vertices and directly dependent subgoals are connected by edges, then dependence is equivalent to connectivity and indivisible subsets are equivalent to connected components of size greater than 1. The divisibility partition is the partition of a graph into connected components, with all the "lonely" vertices collected together, in a special component. Figure 3 shows an example of such a graph for the set $\mathcal{S}_0$ and for an empty binding set. The whole set is divisible into four mutually independent subsets. The subsets $\{e(Z), f(Z,V)\}$ and $\{b(X), c(Y), d(X,Y)\}$ are indivisible. Elements of the divisibility partition $DPart(\mathcal{S}_0, \emptyset)$ are shown by dotted lines.

If a subgoal is independent of the set, then its average cost and number of solutions do not depend on its position within the ordered sequence:

$$\overline{cost}(A)|_{\mathcal{B}} = \frac{\sum_{b \in Sols(\mathcal{B})} Cost(A|_b)}{|Sols(\mathcal{B})|} = \frac{|Sols(\mathcal{B})| \times Cost(A)}{|Sols(\mathcal{B})|} = Cost(A),$$

$$\overline{nsols}(A)|_{\mathcal{B}} = \frac{\sum_{b \in Sols(\mathcal{B})} |Sols(\{A|_b\})|}{|Sols(\mathcal{B})|} = \frac{|Sols(\mathcal{B})| \times |Sols(\{A\})|}{|Sols(\mathcal{B})|} = |Sols(\{A\})|.$$

In this case we can omit the binding information and write $\overline{cost}(A_i)$ instead of $\overline{cost}(A_i)|_{\{A_1 \ldots A_{i-1}\}}$, and $\overline{nsols}(A_i)$ instead of $\overline{nsols}(A_i)|_{\{A_1 \ldots A_{i-1}\}}$.

In practice, program rule bodies rarely feature independent sets of literals. An example is the following clause, which states that children like candy:

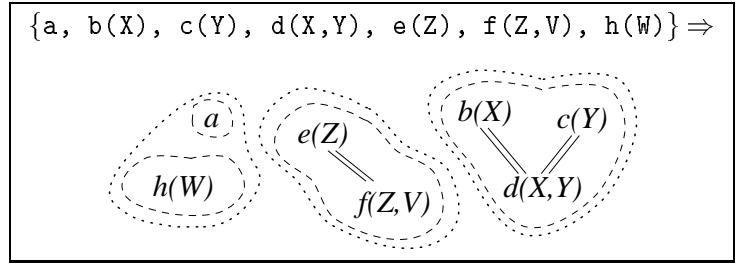

Figure 3: An example of a graph representing a set of subgoals. Directly dependent subgoals are connected by edges. Independent subgoals and indivisible subsets are equivalent to connected components (surrounded by dashed lines). The divisibility partition (under the empty binding set) is shown by dotted lines.

$$\texttt{likes(X,Y)} \leftarrow \texttt{child(X), candy(Y)}.$$

More often, independent rule bodies appear not because they are written as such in the program text, but because some variables are bound in (initially dependent) rule bodies, as a result of clause head unification. For example, if the rule

$$\texttt{father(X,Y)} \leftarrow \texttt{male(X), parent(X,Y)}.$$

is used to reduce `father(abraham,W)`, then `X` is bound to `abraham`, and the rule body becomes independent. Rule bodies often become independent after substitutions are performed in the course of the inference process.

### 3.2.2 Algorithm for Ordering Independent Sets by Sorting

Let $\vec{S}$ be an ordered sub-sequence of subgoals, $\mathcal{B}$ a set of subgoals. We denote

$$cn(\vec{S})|_{\mathcal{B}} = \frac{n\overline{sol}s(\vec{S})|_{\mathcal{B}} - 1}{c\overline{os}t(\vec{S})|_{\mathcal{B}}}.$$

The name "$cn$" reflects the participation of $c\overline{os}t$ and $n\overline{sol}s$ in the definition. When the sub-sequence $\vec{S}$ is independent of other subgoals, the binding information ($|_{\mathcal{B}}$) can be omitted. Together, the average cost, average number of solutions, and $cn$ value of a subgoal will be called *the control values* of this subgoal.

For independent sets, there exists an efficient ordering algorithm, listed in Figure 4. The complexity of this algorithm is $O(n(\tau + \log n))$: $O(n \cdot \tau)$ to obtain the control values of $n$ subgoals, and $O(n \log n)$ to perform the sorting (Knuth, 1973). To enable the division, we must define the cost so that $c\overline{os}t(A_i)$ is always positive. If we define the cost as the number of unifications performed, then always $c\overline{os}t(A_i) \geq 1$, under a reasonable assumption that predicates of all rule body subgoals are defined in the program. (In this case, at least one unification is performed for each subgoal). Similar algorithms were proposed by Simon and Kadane (1975) and Natarajan (1987).

**Example 2** *Let the set of independent subgoals be $\{p, q, r\}$, with the following control values:*

---

**Algorithm 1**

  Let $\mathcal{S} = \{A_1, A_2, \ldots A_n\}$ be a set of subgoals.

  Sort $\mathcal{S}$ using $cn(A_i) = \frac{\overline{\text{nsols}}(A_i) - 1}{\overline{\text{cost}}(A_i)}$ as the key for $A_i$, and return the result.

---

Figure 4: The algorithm for ordering subgoals by sorting.

|  | $p$ | $q$ | $r$ |
|---|---|---|---|
| $\overline{\text{cost}}$ | 10 | 20 | 5 |
| $\overline{\text{nsols}}$ | 1 | 5 | 0.1 |
| cn | 0 | 0.2 | $-0.18$ |

*We compute the costs of all possible orderings, using Equation 5:*

$$
\begin{aligned}
Cost(\langle p, q, r \rangle) &= 10 + 1 \cdot 20 + 1 \cdot 5 \cdot 5 = 55 \\
Cost(\langle p, r, q \rangle) &= 10 + 1 \cdot 5 + 1 \cdot 0.1 \cdot 20 = 17 \\
Cost(\langle q, p, r \rangle) &= 20 + 5 \cdot 10 + 5 \cdot 1 \cdot 5 = 95 \\
Cost(\langle q, r, p \rangle) &= 20 + 5 \cdot 5 + 5 \cdot 0.1 \cdot 10 = 50 \\
Cost(\langle r, p, q \rangle) &= 5 + 0.1 \cdot 10 + 0.1 \cdot 1 \cdot 20 = 8 \\
Cost(\langle r, q, p \rangle) &= 5 + 0.1 \cdot 20 + 0.1 \cdot 5 \cdot 10 = 12
\end{aligned}
$$

*The minimal ordering is $\langle r, p, q \rangle$, and this is exactly the ordering which is found much more quickly by Algorithm 1 for the set $\{p, q, r\}$: $r$ has the smallest $cn$ value, $-0.18$, then goes $p$ with $cn(p) = 0$, and finally $q$ with $cn(q) = 0.2$.*

Note that the sorting algorithm reflects a well-known principle: The best implementations of generate-and-test programs are obtained with the tests placed as early as possible in the rule body and the generations as late as possible (Naish, 1985a). Of course, the cheap tests should come first, while the expensive ones should come last. If one looks at the $cn$ measure, one quickly realizes that tests should be put in front (because $\overline{\text{nsols}} < 1$, so $cn < 0$), while generator subgoals should move towards the end ($\overline{\text{nsols}} > 1$, so $cn > 0$). The weakness of the "test-first" principle is in the fact that not every subgoal can be easily tagged as a test or a generator. If one subgoal has $\overline{\text{nsols}} < 1$ and another one has $\overline{\text{nsols}} > 1$, then their order is obvious even without looking at the costs (because their $cn$ values have different signs). But if both subgoals have $\overline{\text{nsols}} < 1$, or both have $\overline{\text{nsols}} > 1$, then the decision is not so simple. Sorting by $cn$ can correctly handle all the possible cases.

### 3.2.3 CORRECTNESS PROOF OF THE SORTING ALGORITHM FOR INDEPENDENT SETS

We saw that Algorithm 1 found a minimal ordering in Example 2. We are now going to prove that Algorithm 1 *always* finds a minimal ordering for independent sets. First we show an important lemma which will also be used in further discussion. This lemma states that substitution of a sub-sequence by its cheaper permutation makes the entire sequence cheaper.

**Lemma 1**
Let $\vec{S} = \vec{A}\|\vec{B}\|\vec{C}$, $\vec{S}' = \vec{A}\|\vec{B}'\|\vec{C}$, where $\vec{B}$ and $\vec{B}'$ are permutations of one another, and $\vec{A}$ either is empty or has $\mathrm{n\overline{so}ls}(\vec{A}) > 0$. Then

$$Cost(\vec{S}) < Cost(\vec{S}') \iff \mathrm{c\overline{os}t}(\vec{B})|_{\vec{A}} < \mathrm{c\overline{os}t}(\vec{B}')|_{\vec{A}},$$

$$Cost(\vec{S}) = Cost(\vec{S}') \iff \mathrm{c\overline{os}t}(\vec{B})|_{\vec{A}} = \mathrm{c\overline{os}t}(\vec{B}')|_{\vec{A}}.$$

**Proof:** If $\vec{A}$ and $\vec{C}$ are not empty,

$$
\begin{aligned}
Cost(\vec{S}) - Cost(\vec{S}') &= Cost(\vec{A}\|\vec{B}\|\vec{C}) - Cost(\vec{A}\|\vec{B}'\|\vec{C}) = \\
&\overset{(5)}{=} \left( c\overline{os}t(\vec{A})|_\emptyset + n\overline{so}ls(\vec{A})|_\emptyset \times c\overline{os}t(\vec{B})|_{\vec{A}} + n\overline{so}ls(\vec{A}\|\vec{B})|_\emptyset \times c\overline{os}t(\vec{C})|_{\vec{A}\|\vec{B}} \right) - \\
&\quad \left( c\overline{os}t(\vec{A})|_\emptyset + n\overline{so}ls(\vec{A})|_\emptyset \times c\overline{os}t(\vec{B}')|_{\vec{A}} + n\overline{so}ls(\vec{A}\|\vec{B}')|_\emptyset \times c\overline{os}t(\vec{C})|_{\vec{A}\|\vec{B}'} \right).
\end{aligned}
$$

By Theorem 1, $\vec{B}$ and $\vec{B}'$ produce the same solution sets. Hence, the third terms in the parentheses above are equal, and

$$Cost(\vec{S}) - Cost(\vec{S}') = n\overline{so}ls(\vec{A})|_\emptyset \times \left( c\overline{os}t(\vec{B})|_{\vec{A}} - c\overline{os}t(\vec{B}')|_{\vec{A}} \right).$$

Since $n\overline{so}ls(\vec{A}) > 0$, the sign of $Cost(\vec{S}) - Cost(\vec{S}')$ coincides with the sign of $c\overline{os}t(\vec{B})|_{\vec{A}} - c\overline{os}t(\vec{B}')|_{\vec{A}}$.

If $\vec{A}$ or $\vec{C}$ is empty, the proof is similar. □

**Definition:** Let $\vec{S} = \vec{A}\|\vec{B}_1\|\vec{C}\|\vec{B}_2\|\vec{D}$ be an ordered sequence of subgoals ($\vec{A}$, $\vec{C}$ and $\vec{D}$ may be empty sequences). With respect to $\vec{S}$, the pair $\langle \vec{B}_1, \vec{B}_2 \rangle$ is

- *cn-ordered*, if $cn(\vec{B}_1)|_{\vec{A}} \le cn(\vec{B}_2)|_{\vec{A} \cup \vec{B}_1 \cup \vec{C}}$

- *cn-inverted*, if $cn(\vec{B}_1)|_{\vec{A}} > cn(\vec{B}_2)|_{\vec{A} \cup \vec{B}_1 \cup \vec{C}}$

We now show that two adjacent *mutually independent* sequences of subgoals in a minimal ordering must be cn-ordered.

**Lemma 2**
Let $\vec{S} = \vec{A}\|\vec{B}_1\|\vec{B}_2\|\vec{C}$, $\vec{S}' = \vec{A}\|\vec{B}_2\|\vec{B}_1\|\vec{C}$, where $\vec{B}_1$, $\vec{B}_2$ are mutually independent under $\vec{A}$. Let $\vec{A}$ either be empty or have $\mathrm{n\overline{so}ls}(\vec{A}) > 0$. Then

$$
\begin{aligned}
Cost(\vec{S}) < Cost(\vec{S}') &\iff cn(\vec{B}_1)|_{\vec{A}} < cn(\vec{B}_2)|_{\vec{A}}, \\
Cost(\vec{S}) = Cost(\vec{S}') &\iff cn(\vec{B}_1)|_{\vec{A}} = cn(\vec{B}_2)|_{\vec{A}}.
\end{aligned}
$$

**Proof:**

$$Cost(\vec{S}) < Cost(\vec{S}') \overset{\text{Lemma 1}}{\iff} c\overline{os}t(\vec{B}_1 \| \vec{B}_2)|_{\vec{A}} < c\overline{os}t(\vec{B}_2 \| \vec{B}_1)|_{\vec{A}}$$

$$\iff c\overline{os}t(\vec{B}_1)|_{\vec{A}} + n\overline{so}ls(\vec{B}_1)|_{\vec{A}} \times c\overline{os}t(\vec{B}_2)|_{\vec{A} \cup \vec{B}_1} <$$
$$c\overline{os}t(\vec{B}_2)|_{\vec{A}} + n\overline{so}ls(\vec{B}_2)|_{\vec{A}} \times c\overline{os}t(\vec{B}_1)|_{\vec{A} \cup \vec{B}_2}$$

$$\overset{\text{indep.}\{\vec{B}_1,\vec{B}_2\}}{\iff} c\overline{os}t(\vec{B}_1)|_{\vec{A}} + n\overline{so}ls(\vec{B}_1)|_{\vec{A}} \times c\overline{os}t(\vec{B}_2)|_{\vec{A}} <$$
$$c\overline{os}t(\vec{B}_2)|_{\vec{A}} + n\overline{so}ls(\vec{B}_2)|_{\vec{A}} \times c\overline{os}t(\vec{B}_1)|_{\vec{A}}$$

$$\iff n\overline{so}ls(\vec{B}_1)|_{\vec{A}} \times c\overline{os}t(\vec{B}_2)|_{\vec{A}} - c\overline{os}t(\vec{B}_2)|_{\vec{A}} <$$
$$n\overline{so}ls(\vec{B}_2)|_{\vec{A}} \times c\overline{os}t(\vec{B}_1)|_{\vec{A}} - c\overline{os}t(\vec{B}_1)|_{\vec{A}}$$

$$\overset{c\overline{os}t(\vec{B}_i)|_{\vec{A}} > 0}{\iff} \frac{n\overline{so}ls(\vec{B}_1)|_{\vec{A}} - 1}{c\overline{os}t(\vec{B}_1)|_{\vec{A}}} < \frac{n\overline{so}ls(\vec{B}_2)|_{\vec{A}} - 1}{c\overline{os}t(\vec{B}_2)|_{\vec{A}}}$$

$$\iff cn(\vec{B}_1)|_{\vec{A}} < cn(\vec{B}_2)|_{\vec{A}}$$

$$Cost(\vec{S}) = Cost(\vec{S}') \iff cn(\vec{B}_1)|_{\vec{A}} = cn(\vec{B}_2)|_{\vec{A}} \text{ — similar.} \qquad \square$$

In an *independent set*, all subgoal pairs are independent, in particular all adjacent pairs. So, in a minimal ordering of an independent set, all adjacent subgoal pairs must be cn-ordered; otherwise, the cost of the sequence can be reduced by a transposition of such pair. This conclusion is expressed in the following theorem.

**Theorem 2**
*Let $\mathcal{S}$ be an independent set. Let $\vec{S}$ be an ordering of $\mathcal{S}$. $\vec{S}$ is minimal iff all the subgoals in $\vec{S}$ are sorted in non-decreasing order by their cn values.*

**Proof:**

1. Let $\vec{S}$ be a minimal ordering of $\mathcal{S}$. If $\vec{S}$ contains a cn-inverted adjacent pair of subgoals, then transposition of this pair reduces the cost of $\vec{S}$ (Lemma 2), contradicting the minimality of $\vec{S}$.

2. Let $\vec{S}$ be some ordering of $\mathcal{S}$, whose subgoals are sorted in non-decreasing order by $cn$. Let $\vec{S}'$ be a minimal ordering of $\mathcal{S}$. According to item 1, $\vec{S}'$ is also sorted by $cn$. The only possible difference between the two sequences is the internal ordering of sub-sequences with equal $cn$ values. The ordering of each such sub-sequence in $\vec{S}$ can be transformed to the ordering of its counterpart sub-sequence in $\vec{S}'$ by a finite number of transpositions of adjacent subgoals. By Lemma 2, transpositions of adjacent independent subgoals with equal $cn$ values cannot change the cost of the sequence. Therefore, $Cost(\vec{S}) = Cost(\vec{S}')$, and $\vec{S}$ is a minimal ordering of $\mathcal{S}$ (since $\vec{S}'$ is minimal). $\qquad \square$

**Corollary 1** *Algorithm 1 finds a minimal ordering of an independent set of subgoals.*

### 3.3 Ordering of Dependent Sets of Subgoals

Algorithm 1 does not guarantee finding a minimal ordering when the given set of subgoals is dependent, as the following proposition shows.

**Proposition 1** *When the given set of subgoals is dependent, then:*

1. *The result of Algorithm 1 on it is not always defined.*

2. *Even when the result is defined, it is not always a minimal ordering of the set.*

**Proof:** Both claims are proved by counter-examples.

1. We show a set of subgoals that cannot be ordered by sorting.

   The program:

   ```
   a(c1).   b(c1).
   a(c2).   b(c2).
   ```

   Control values:

   |  | $a(X)\|_\emptyset$ | $a(X)\|_{\{b(X)\}}$ | $b(X)\|_\emptyset$ | $b(X)\|_{\{a(X)\}}$ |
   |---|---|---|---|---|
   | $c\overline{os}t$ | 2 | 2 | 2 | 2 |
   | $n\overline{so}ls$ | 2 | 1 | 2 | 1 |
   | $cn$ | $\frac{1}{2}$ | 0 | $\frac{1}{2}$ | 0 |

   The set {a(X), b(X)} has two possible orderings, $\langle a(X), b(X) \rangle$ and $\langle b(X), a(X) \rangle$. Both orderings have minimal cost, though neither one is sorted by $cn$: each ordering has $cn = \frac{1}{2}$ for its first subgoal, and $cn = 0$ for the second one. Sorting by $cn$ is impossible here: when we transpose subgoals, their $cn$ values are changed, and the pair becomes cn-inverted again.

2. We show a set of subgoals that can be ordered by sorting, but its sorted ordering is not minimal.

   The program:

   ```
   a(c1).
   a(c1).
   b(c1).
   b(c2) ← a(c1), a(c2).
   ```

   Control values:

   |  | $a(X)\|_\emptyset$ | $a(X)\|_{\{b(X)\}}$ | $b(X)\|_\emptyset$ | $b(X)\|_{\{a(X)\}}$ |
   |---|---|---|---|---|
   | $c\overline{os}t$ | 2 | 2 | 8 | 2 |
   | $n\overline{so}ls$ | 2 | 2 | 1 | 1 |
   | $cn$ | $\frac{1}{2}$ | $\frac{1}{2}$ | 0 | 0 |

   Let the unordered set of subgoals be {a(X), b(X)}. Its ordering $\langle b(X), a(X) \rangle$ is sorted by $cn$, while $\langle a(X), b(X) \rangle$ is not. But $\langle a(X), b(X) \rangle$ is cheaper than $\langle b(X), a(X) \rangle$:

   $$c\overline{os}t(\langle a(X), b(X) \rangle) = 2 + 2 \cdot 2 = 6 \qquad c\overline{os}t(\langle b(X), a(X) \rangle) = 8 + 1 \cdot 2 = 10$$

   $\square$

Since sorting cannot guarantee minimal ordering for dependent subgoals, we now consider alternative ordering algorithms. The simplest algorithm checks every possible permutation of the set and returns the one with the minimal cost. The listing for this algorithm is shown in Figure 5.

This algorithm runs in $O(\tau \cdot n!)$ time, where $\tau$ is the time it takes to compute the control values for one subgoal, and $n$ is the number of subgoals.

The following observation can help to reduce the ordering time at the expense of additional space. Ordered sequences can be constructed incrementally, by adding subgoals to

---

**Algorithm 2**

*For each permutation of subgoals, find its cost according to Equation 5.*
*Store the currently cheapest permutation and update it when a cheaper*
*one is found.*
*Finally, return the cheapest permutation.*

---

Figure 5: The algorithm for subgoal ordering by an exhaustive check of all permutations.

---

**Algorithm 3**

**Order($\mathcal{S}$)**
    let $\mathcal{P}_0 \leftarrow \{\emptyset\}$, $n \leftarrow |\mathcal{S}|$
    loop for $k = 1$ to $n$
        $\mathcal{P}'_k \leftarrow \left\{ \vec{P} \| B \;\middle|\; \vec{P} \in \mathcal{P}_{k-1}, \; B \in \mathcal{S} \setminus \vec{P} \right\}$
        $\mathcal{P}_k \leftarrow \left\{ \vec{P} \in \mathcal{P}'_k \;\middle|\; \forall \vec{P}' \in \mathcal{P}'_k, \; \left[ permutation(\vec{P}, \vec{P}') \Rightarrow Cost(\vec{P}) \le Cost(\vec{P}') \right] \right\}$
    Return *the single member of* $\mathcal{P}_n$.

---

Figure 6: The ordering algorithm which checks permutations of ordered prefixes.

the right ends of ordered prefixes. By Lemma 1, if a cheaper permutation of a prefix exists, then this prefix cannot belong to a minimal ordering. The ordering algorithm can build prefixes with increasing lengths, at each step adding to the right end of each prefix one of the subgoals that do not appear in it already, and for each subset keeping only its cheapest permutation (if several permutations have equal cost, any one of them can be chosen). The listing for this algorithm is shown in Figure 6. At each step $k$, $\mathcal{P}'_k$ stores the set of prefixes from step $k - 1$ extended by every subgoal not appearing there already. $\mathcal{P}_k \subseteq \mathcal{P}'_k$, and in $\mathcal{P}_k$ each subset of subgoals is represented only by its cheapest permutation. Obviously, $|\mathcal{P}_k| = \binom{n}{k}$ (one prefix is kept for every subset of $\mathcal{S}$ of size $k$). For each prefix of length $k - 1$, there are $n - (k - 1)$ possible continuations of length $k$. The size of $\mathcal{P}'_k$ is as follows:

$$|\mathcal{P}'_k| = \binom{n}{k-1} \cdot (n - (k-1)) = \frac{n!}{(n - (k-1))!(k-1)!} \cdot (n - (k-1)) = \frac{k}{k} \cdot \frac{n!}{(n-k)!(k-1)!} = k \cdot \binom{n}{k}.$$

For each prefix, we compute its cost in $\tau$ time. The permutation test can be completed in $O(n)$ time, by using, for example, a trie structure (Aho et al., 1987), where subgoals in prefixes are sorted lexicographically. Each step $k$ takes $O((n + \tau) \cdot k \cdot \binom{n}{k})$ time, and the

whole algorithm runs in

$$\sum_{k=1}^{n} O((n+\tau) \cdot k \cdot \binom{n}{k}) = O(n \cdot (n+\tau) \cdot \sum_{k=1}^{n} \binom{n}{k}) = O(n \cdot (n+\tau) \cdot 2^n).$$

If $\tau = O(n)$, this makes $O(n^2 \cdot 2^n)$.

Smith and Genesereth (1985) and Natarajan (1987) point out that in a minimal ordered sequence every adjacent pair of subgoals must satisfy an *adjacency restriction*. The most general form of such a restriction in our notation says that two adjacent subgoals $A_k$ and $A_{k+1}$ in a minimal ordering $\langle A_1, A_2 \ldots A_n \rangle$ must satisfy

$$c\overline{os}t(\langle A_k, A_{k+1} \rangle)|_{\{A_1 \ldots A_{k-1}\}} \leq c\overline{os}t(\langle A_{k+1}, A_k \rangle)|_{\{A_1 \ldots A_{k-1}\}}. \tag{8}$$

The restriction follows immediately from Lemma 1. However, it can only help to find a *locally* minimal ordering, i.e., an ordering that cannot be improved by transpositions of adjacent subgoals. It is possible that all adjacent subgoal pairs satisfy Equation 8, but the ordering is still not minimal. The following example illustrates this statement.

**Example 3** *Let the unordered set be* $\{p(X), q(X), r(X)\}$, *where the predicates are defined by the following program:*

$$\begin{array}{lll}
p(c_1). & q(c_1). & r(c_1). \\
p(c_2) \leftarrow f. & q(c_2). & r(c_1). \\
 & q(c_3) \leftarrow f. & \\
f \leftarrow \textit{fails after 50 unifications.} &
\end{array}$$

*The ordering* $\langle p(X), q(X), r(X) \rangle$ *satisfies the adjacency restriction (Equation 8):*

$$\begin{array}{ll}
c\overline{os}t(p(X), q(X))|_\emptyset = 55 & c\overline{os}t(q(X), r(X))|_{p(X)} = 5 \\
c\overline{os}t(q(X), p(X))|_\emptyset = 107 & c\overline{os}t(r(X), q(X))|_{p(X)} = 8
\end{array}$$

*But it is not minimal:*

$$\begin{array}{l}
c\overline{os}t(\langle p(X), q(X), r(X) \rangle) = 57 \\
c\overline{os}t(\langle r(X), p(X), q(X) \rangle) = 12
\end{array}$$

To find a globally minimal ordering, it seems beneficial to combine the prefix algorithm with the adjacency restriction: if a prefix does not satisfy the adjacency restriction, then there is a cheaper permutation of this prefix. The adjacency test can be performed faster than the permutation test, since it must only consider the two last subgoals of each prefix. Nevertheless, the number of prefixes remaining after each step of Algorithm 3 is not reduced: if a prefix is rejected due to a violation of the adjacency restriction, it would have also been rejected by the permutation test. Furthermore, if the adjacency restriction test does not fail, we should still perform the permutation test to avoid local minima (as in Example 3). The adjacency test succeeds in at least half of the cases: if we examine a prefix $\langle A_1, \ldots A_k, B_1, B_2 \rangle$, we shall also examine $\langle A_1, \ldots A_k, B_2, B_1 \rangle$, and the adjacency test cannot fail in both. Consequently, addition of the adjacency test can only halve the total running time of the ordering algorithm, leaving it $O(n^2 \cdot 2^n)$ in the worst case.

Smith and Genesereth propose performing a best-first search in the space of ordered prefixes, preferring prefixes with lower cost. The best-first search can be combined with the permutation test and the adjacency restriction. In addition, when the subgoals not in a prefix are independent under its binding, they can be sorted, and the sorted result concatenated to the prefix. By Lemma 1 and Corollary 1, this produces the cheapest completion of this prefix. When we perform completion, there is no need to perform the adjacency or permutation test: if a complete sequence is not minimal, it will never be chosen as the cheapest prefix; even if it is added to the list of prefixes, it will never be extracted therefrom. The resulting algorithm is shown in Figure 7.

---

**Algorithm 4**

    **Order**($\mathcal{S}$)
        let *prefix-list* $\leftarrow \emptyset$, *prefix* $\leftarrow \emptyset$, *rest* $\leftarrow \mathcal{S}$
        loop until **empty**(*rest*)
            if **Independent**($rest|_{prefix}$)
            then
                let *completion* $\leftarrow$ *prefix*$\|$**Sort-by-cn**($rest|_{prefix}$)
                **Insert-By-Cost**(*completion*, *prefix-list*)
            else
                loop for *subgoal* $\in$ *rest*
                    let *extension* $\leftarrow$ *prefix*$\|$*subgoal*
                    if **Adjacency-Restriction-Test**(*extension*)
                    and **Permutation-Test**(*extension*)
                    then
                        **Insert-By-Cost**(*extension*, *prefix-list*)
            *prefix* $\leftarrow$ **Cheapest**(*prefix-list*)
            **Remove-from-list**(*prefix*, *prefix-list*)
            *rest* $\leftarrow \mathcal{S}\backslash prefix$
        Return *prefix*

---

Figure 7: An algorithm for subgoal ordering, incorporating the ideas of earlier researchers.

The advantage of using best-first search is that it avoids expanding prefixes whose cost is higher than the cost of the minimal ordering. The policy used by the algorithm may, however, be suboptimal or even harmful. It often happens that the best completion of a cheaper prefix is much more expensive than the best completion of a more expensive prefix. When the number of solutions is large, it is better to place subgoals with high costs closer to the beginning of the ordering to reduce the number of times that their cost is multiplied.

For example, let the set be $\{a(X), b(X)\}$, with $\overline{cost}(a(X)) = 10$, $\overline{cost}(b(X)) = n\overline{sols}(a(X)) = n\overline{sols}(b(X)) = 2$. Then a minimal ordering starts with the most expensive prefix:

$$Cost(\langle a(X), b(X) \rangle) = 10 + 2 \cdot 2 = 14$$

$$Cost(\langle b(X), a(X)\rangle) = 2 + 2 \cdot 10 = 22$$

If there are many prefixes whose cost is higher than the cost of the minimal ordering, then best-first search saves time. But if the number of such prefixes is small, using best-first search can increase the total time, due to the need to perform insertion of a prefix into a priority queue, according to its cost.

A sample run of Algorithm 4 will be shown later (in Section 4.7).

## 4. The Divide-and-Conquer Subgoal Ordering Algorithm

Algorithm 1 presented in Section 3.2 is very efficient, but is applicable only when the entire set of subgoals is independent. Algorithm 3 can handle a dependent set of subgoals but is very inefficient. Algorithm 4, a combination of the two, can exploit independence of subgoals for better efficiency. However, the obtained benefit is quite limited. In this section, we present the Divide-and-Conquer (DAC) algorithm, which is able to exploit subgoal independence in a more elaborate way. The algorithm divides the set of subgoals into smaller subsets, orders these subsets recursively and combines the results.

### 4.1 Divisibility Trees of Subgoal Sets

In this subsection we define a structure that represents all the ways of breaking a subgoal set into independent parts. Our algorithm will work by traversing this structure.

**Definition:** Let $\mathcal{S}$ and $\mathcal{B}$ be sets of subgoals. The *divisibility tree of $\mathcal{S}$ under $\mathcal{B}$*, $DTree(\mathcal{S}, \mathcal{B})$, is an AND-OR tree defined as follows:

$$DTree(\mathcal{S}, \mathcal{B}) = \begin{cases} \text{leaf}(\mathcal{S}, \mathcal{B}) & - \; \mathcal{S} \text{ is independent under } \mathcal{B} \\ \text{OR}(\mathcal{S}, \mathcal{B}, \{DTree(\mathcal{S} \setminus \{B_i\}, \mathcal{B} \cup \{B_i\}) \mid B_i \in \mathcal{S}\}) & - \; \mathcal{S} \text{ is indivisible under } \mathcal{B} \\ \text{AND}(\mathcal{S}, \mathcal{B}, \{DTree(\mathcal{S}_i, \mathcal{B}) \mid \mathcal{S}_i \in DPart(\mathcal{S}, \mathcal{B})\}) & - \; \mathcal{S} \text{ is divisible under } \mathcal{B} \end{cases}$$

Each node $N$ in the tree $DTree(\mathcal{S}_0, \mathcal{B}_0)$ has an associated set of subgoals $\mathcal{S}(N) \subseteq \mathcal{S}_0$ and an associated binding set $\mathcal{B}(N) \supseteq \mathcal{B}_0$. For the root node, $\mathcal{S}(N) = \mathcal{S}_0$, $\mathcal{B}(N) = \mathcal{B}_0$. If the binding set of the root is not specified explicitly, we assume it to be empty. For AND-nodes and OR-nodes we also define the sets of children.

- If $\mathcal{S}(N)$ is *independent* under $\mathcal{B}(N)$, then $N$ is a leaf.

- If $\mathcal{S}(N)$ is *indivisible* under $\mathcal{B}(N)$, then $N$ is an OR-node. Each subgoal $B_i$ in $\mathcal{S}(N)$ defines a child node whose set of subgoals is $\mathcal{S}(N) \setminus \{B_i\}$ and the binding set is $\mathcal{B}(N) \cup \{B_i\}$. We call $B_i$ the *binder* of the generated child. Note that the binding set of every node in a divisibility tree is the union of the binders of all its indivisible ancestors and of the root's binding set.

- If $\mathcal{S}(N)$ is *divisible* under $\mathcal{B}(N)$, then $N$ is an AND-node. Each subset $\mathcal{S}_i$ in the divisibility partition $DPart(\mathcal{S}(N), \mathcal{B}(N))$ defines a child node with associated set of subgoals $\mathcal{S}_i$ and binding set $\mathcal{B}(N)$. Divisibility partition was defined in Section 3.2.1.

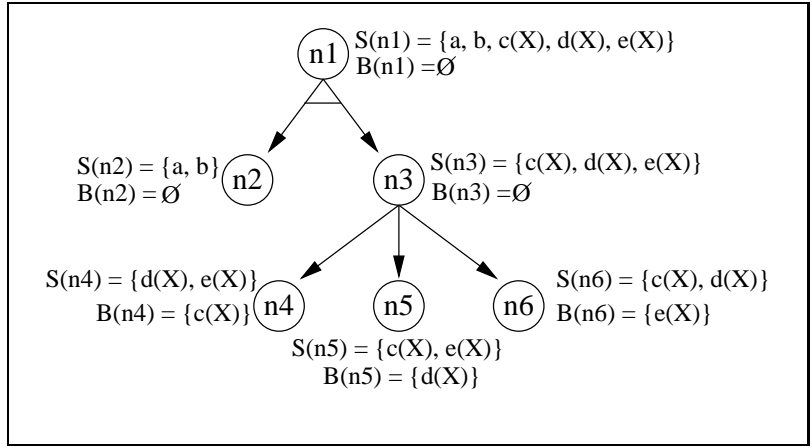

Figure 8: The divisibility tree of $\{a, b, c(X), d(X), e(X)\}$ under empty initial binding set. The set associated with node $n1$ is divisible, and is represented by an AND-node. Its children correspond to its divisibility subsets – one independent, $S(n2) = \{a, b\}$, and one indivisible, $S(n3) = \{c(X), d(X), e(X)\}$. $n3$ is an OR-node, whose children correspond to its three subgoals (each subgoal serves as a binder in one of the children). The sets $S(n2)$, $S(n4)$, $S(n5)$ and $S(n6)$ are independent under their respective binding sets, and their nodes are leaves. Here we assumed that the subgoals $c(X)$, $d(X)$ and $e(X)$ bind $X$ as a result of their proof.

It is easy to show that the divisibility tree of a set of subgoals is unique up to the order of children of each node. Figure 8 shows the divisibility tree of the set $\{a, b, c(X), d(X), e(X)\}$ under empty initial binding set. The associated sets and binding sets are written next to the nodes.

The following lemma expresses an important property of divisibility trees: subgoals of each node are independent of the rest of subgoals under the binding set of the node.

**Lemma 3** *Let $\mathcal{S}_0$ be a set of subgoals. Then for every node $N$ in $DTree(\mathcal{S}_0, \emptyset)$, for every subgoal $A \in \mathcal{S}(N)$, and for every subgoal $Y \in \mathcal{S}_0 \setminus (\mathcal{S}(N) \cup \mathcal{B}(N))$, $A$ and $Y$ are independent under $\mathcal{B}(N)$.*

**Proof:** by induction on the depth of $N$ in the divisibility tree.

**Inductive base:** $N$ is the root node, $\mathcal{S}_0 \setminus \mathcal{S}(N)$ is empty, and no such $Y$ exists.

**Inductive hypothesis:** The lemma holds for $M$, the parent node of $N$.

**Inductive step:** Let $A \in \mathcal{S}(N)$, $Y \in \mathcal{S}_0 \setminus (\mathcal{S}(N) \cup \mathcal{B}(N))$. $A \in \mathcal{S}(M)$, and for $M$ the lemma holds, thus either $A$ and $Y$ are independent under $\mathcal{B}(M)$, or $Y \in \mathcal{S}(M)$.

If $A$ and $Y$ are independent under $\mathcal{B}(M)$, then they are also independent under $\mathcal{B}(N)$, since $\mathcal{B}(M) \subseteq \mathcal{B}(N)$. Otherwise, $A$ and $Y$ are dependent under $\mathcal{B}(M)$, and $Y \in \mathcal{S}(M)$.

- If $M$ is an AND-node, and $A$ and $Y$ are dependent under $\mathcal{B}(M)$, then $A$ and $Y$ belong to the same element of $DPart(\mathcal{S}(M), \mathcal{B}(M))$, and $Y \in \mathcal{S}(N)$ – a contradiction.

- If $M$ is an OR-node and $Y \in \mathcal{S}(M) \setminus \mathcal{S}(N)$, then $Y$ must be the binder of $N$. But then $\mathcal{B}(N) = \mathcal{B}(M) \cup \{Y\}$ and $Y \in \mathcal{B}(N)$ – a contradiction again. $\qquad\square$

The lemma relates to subgoal independence inside divisibility trees. We shall sometimes need to argue about independence inside ordered sequences of subgoals. The following corollary provides the necessary connecting link.

**Corollary 2** *Let $\mathcal{S}_0$ be a set of subgoals, $N$ be a node in the divisibility tree of $\mathcal{S}_0$, $\vec{S}$ an ordering of $\mathcal{S}_0$, $\vec{S} = \vec{S}_1 \| \vec{S}_2$, where $\mathcal{B}(N) \subseteq \vec{S}_1$ and $\mathcal{S}(N) \subseteq \vec{S}_2$. Then $\mathcal{S}(N)$ is mutually independent of $\vec{S}_2 \setminus \mathcal{S}(N)$ under $\vec{S}_1$.*

**Proof:** Let $A \in \mathcal{S}(N)$, $Y \in \vec{S}_2 \setminus \mathcal{S}(N)$. $A$ and $Y$ are independent under $\mathcal{B}(N)$, by the preceding lemma. Since $\mathcal{B}(N) \subseteq \vec{S}_1$, $A$ and $Y$ are independent under $\vec{S}_1$. Every subgoal of $\mathcal{S}(N)$ is independent of every subgoal of $\vec{S}_2 \setminus \mathcal{S}(N)$ under $\vec{S}_1$; therefore, $\mathcal{S}(N)$ and $\vec{S}_2 \setminus \mathcal{S}(N)$ are mutually independent under $\vec{S}_1$. $\qquad\square$

## 4.2 Valid Orderings in Divisibility Trees

The aim of our ordering algorithm is to find a minimal ordering of a given set of subgoals. We construct orderings following a divide-and-conquer policy: larger sets are split into smaller ones, and orderings of the smaller sets are combined to produce an ordering of the larger set. To implement this policy, we perform a post-order traversal of the divisibility tree corresponding to the given set of subgoals under an empty initial binding set. When orderings of child nodes are combined to produce an ordering of the parent node, the inner order of their subgoals is not changed: smaller orderings are *consistent* with larger orderings.

**Definition:** Let $\mathcal{S}$ and $\mathcal{G} \subseteq \mathcal{S}$ be sets of subgoals. An ordering $\vec{O}_G$ of $\mathcal{G}$ and an ordering $\vec{O}_S$ of $\mathcal{S}$ are *consistent* (denoted $Cons(\vec{O}_G, \vec{O}_S)$), if the order of subgoals of $\mathcal{G}$ in $\vec{O}_G$ and in $\vec{O}_S$ is the same.

The divide-and-conquer process described above seems analogous to Merge Sort (Knuth, 1973). There, the set of numbers is split into two (or more) subsets, each subset is independently ordered to a sequence consistent with the global order, and these sequences are merged. Is it possible to use a similar method for subgoal ordering? Assume that a set of subgoals is partitioned into two mutually independent subsets, $\mathcal{A}$ and $\mathcal{B}$. Can we build an algorithm that, given $\mathcal{A}$, produces its ordering consistent with a minimal ordering of $\mathcal{A} \cup \mathcal{B}$, independently of $\mathcal{B}$? Unfortunately, the answer is negative. An ordering of $\mathcal{A}$ may be consistent with a minimal ordering of $\mathcal{A} \cup \mathcal{B}_1$ but at the same time not be consistent with a minimal ordering of $\mathcal{A} \cup \mathcal{B}_2$ for some $\mathcal{B}_1 \neq \mathcal{B}_2$.

For example, let $\mathcal{A} = \{a1(X), a2(X)\}$, $\mathcal{B}_1 = \{b\}$, $\mathcal{B}_2 = \{d\}$ and the control values be as specified in Figure 9. The single minimal ordering of $\mathcal{A} \cup \mathcal{B}_1$ is $\langle a2(X), b, a1(X) \rangle$, while the single minimal ordering of $\mathcal{A} \cup \mathcal{B}_2$ is $\langle d, a1(X), a2(X) \rangle$. There is no ordering of $\mathcal{A}$ consistent with both these minimal global orderings.

The program:

```
a1(c1).        b ← a1(X).
a1(c1).        b ← d.
a2(c1).        d.
a2(c1).
a2(c2) ← a1(c2).
```

The control values:

|            | $a1(X)|_\emptyset$ | $a1(X)|_{\{a2(X)\}}$ | $a2(X)|_\emptyset$ | $a2(X)|_{\{a1(X)\}}$ | $b$ | $d$ |
|------------|------|------|------|------|-----|-----|
| $\overline{cost}$ | 2 | 2 | 5 | 3 | 5 | 1 |
| $\overline{nsols}$ | 2 | 2 | 2 | 2 | 3 | 1 |

$Cost(b, a1(X), a2(X)) = 5 + 3 \cdot 2 + 3 \cdot 2 \cdot 3 = 29$         $Cost(d, \underline{a1}(X), \underline{a2}(X)) = 1 + 1 \cdot 2 + 1 \cdot 2 \cdot 3 = \mathbf{9}$

$Cost(b, a2(X), a1(X)) = 5 + 3 \cdot 5 + 3 \cdot 2 \cdot 2 = 32$         $Cost(d, a2(X), a1(X)) = 1 + 1 \cdot 5 + 1 \cdot 2 \cdot 2 = 10$

$Cost(a1(X), b, a2(X)) = 2 + 2 \cdot 5 + 2 \cdot 3 \cdot 3 = 30$         $Cost(a1(X), d, a2(X)) = 2 + 2 \cdot 1 + 2 \cdot 1 \cdot 3 = 10$

$Cost(a1(X), a2(X), b) = 2 + 2 \cdot 3 + 2 \cdot 2 \cdot 5 = 28$         $Cost(a1(X), a2(X), d) = 2 + 2 \cdot 3 + 2 \cdot 2 \cdot 1 = 12$

$Cost(\underline{a2}(X), b, \underline{a1}(X)) = 5 + 2 \cdot 5 + 2 \cdot 3 \cdot 2 = \mathbf{27}$         $Cost(a2(X), d, a1(X)) = 5 + 2 \cdot 1 + 2 \cdot 1 \cdot 2 = 11$

$Cost(a2(X), a1(X), b) = 5 + 2 \cdot 2 + 2 \cdot 2 \cdot 5 = 29$         $Cost(a2(X), a1(X), d) = 5 + 2 \cdot 2 + 2 \cdot 2 \cdot 1 = 13$

Figure 9: We show a small program and the control values it defines. Then we compute costs of all permutations of the sets $\{b, a1(X), a2(X)\}$ and $\{d, a1(X), a2(X)\}$. Different orderings of $\{a1(X), a2(X)\}$ are consistent with minimal orderings of these sets.

Since, unlike the case of Merge Sort, we cannot always identify a single ordering of the subset consistent with a minimal ordering of the whole set, our algorithm will deal with *sets of candidate orderings*. Our requirement from such a set is that it contain at least one local ordering consistent with a global minimal ordering, if such a local ordering exists ("local" ordering is an ordering of the set of the node, "global" ordering is an ordering of the set of the root). Such a set will be called *valid*. The following definition defines valid sets formally, together with several other concepts.

**Definition:** Let $\mathcal{S}_0$ be a set of subgoals and $N$ be a node in the divisibility tree of $\mathcal{S}_0$. Recall that $\pi(\mathcal{S})$ denotes the set of all permutations of $\mathcal{S}$.

1. $\vec{O}_S \in \pi(\mathcal{S}_0)$ is *binder-consistent* with $\vec{O}_N \in \pi(\mathcal{S}(N))$ (denoted $BC_N(\vec{O}_N, \vec{O}_S)$), if they are consistent, and all subgoals of $\mathcal{B}(N)$ appear in $\vec{O}_S$ before all subgoals of $\vec{O}_N$:

$$BC_N(\vec{O}_N, \vec{O}_S) \iff \exists \vec{O}_B \in \pi(\mathcal{B}(N)) : Cons(\vec{O}_B \| \vec{O}_N, \vec{O}_S).$$

$\vec{O}_S \in \pi(\mathcal{S}_0)$ is *binder-consistent* with the node $N$ (denoted $BC_N(\vec{O}_S)$), if it is binder-consistent with some ordering of $\mathcal{S}(N)$:

$$BC_N(\vec{O}_S) \iff \exists \vec{O}_N \in \pi(\mathcal{S}(N)) : BC_N(\vec{O}_N, \vec{O}_S).$$

2. $\vec{O}_N \in \pi(\mathcal{S}(N))$ is *min-consistent* with $\vec{O}_S \in \pi(\mathcal{S}_0)$ (denoted $MC_{N, \mathcal{S}_0}(\vec{O}_N, \vec{O}_S)$), if they are binder-consistent, and $\vec{O}_S$ is minimal:

$$MC_{N, \mathcal{S}_0}(\vec{O}_N, \vec{O}_S) \iff BC_N(\vec{O}_N, \vec{O}_S) \wedge Min(\vec{O}_S, \mathcal{S}_0).$$

$\vec{O}_N \in \pi(\mathcal{S}(N))$ is *min-consistent* (denoted $MC_{N, \mathcal{S}_0}(\vec{O}_N)$), if it is min-consistent with some ordering of $\mathcal{S}_0$:

$$MC_{N, \mathcal{S}_0}(\vec{O}_N) \iff \exists \vec{O}_S \in \pi(\mathcal{S}_0) : MC_{N, \mathcal{S}_0}(\vec{O}_N, \vec{O}_S).$$

3. An ordering $\vec{O}_N \in \pi(\mathcal{S}(N))$ is *MC-contradicting*, if it is not min-consistent:

$$\mathrm{MCC}_{N,\mathcal{S}_0}(\vec{O}_N) \iff \neg MC_{N,\mathcal{S}_0}(\vec{O}_N).$$

4. Two orderings $\vec{O}_1, \vec{O}_2 \in \pi(\mathcal{S}(N))$ are *MC-equivalent*, if one of them is min-consistent iff the other one is:

$$\mathrm{MCE}_{N,\mathcal{S}_0}(\vec{O}_1, \vec{O}_2) \iff [MC_{N,\mathcal{S}_0}(\vec{O}_1) \iff MC_{N,\mathcal{S}_0}(\vec{O}_2)].$$

5. A set of orderings $\mathcal{C}_N \subseteq \pi(\mathcal{S}(N))$ is *valid*, if $\mathcal{C}_N$ contains a min-consistent ordering (when at least one min-consistent ordering of $\mathcal{S}(N)$ exists):

$$Valid_{N,\mathcal{S}_0}(\mathcal{C}_N) \iff [\exists \vec{O}'_N \in \pi(\mathcal{S}(N)) : MC_{N,\mathcal{S}_0}(\vec{O}'_N)] \to [\exists \vec{O}_N \in \mathcal{C}_N : MC_{N,\mathcal{S}_0}(\vec{O}_N)].$$

An important property of valid sets is that a valid set of orderings of the root of $DTree(\mathcal{S}_0, \emptyset)$ must contain a minimal ordering of $\mathcal{S}_0$. Indeed, in the root $\mathcal{S}(N) = \mathcal{S}_0$, and consistency becomes identity. Also, $\mathcal{B}(N) = \emptyset$, so that binder-consistency becomes consistency, and min-consistency becomes minimality. Since there always exists a minimal ordering of $\mathcal{S}_0$, a valid set of orderings of the root must contain a minimal ordering of $\mathcal{S}_0$.

### 4.3 The Outline of the Divide-and-Conquer Algorithm

We propose an algorithm that is based on producing valid sets of orderings. Each node in a divisibility tree produces a valid set for its associated set of subgoals, and passes it to its parent node. After the valid set of the root node is found, we compare costs of all its members, and return the cheapest one.

The set of orderings produced by the algorithm for a node $N$ is called a *candidate set* of $N$. Its members are called *candidate orderings* of $N$, or simply *candidates*. To find a candidate set of $N$, we first consider the set of all possible orderings of $\mathcal{S}(N)$ that are consistent with candidates of $N$'s children. This set is called the *consistency set* of $N$. Given the candidate sets of $N$'s children, the consistency set of $N$ is defined uniquely. A candidate set of $N$ is usually not unique.

**Definition:** Let $N$ be a node in a divisibility tree of $\mathcal{S}_0$. The *consistency set* of $N$, denoted $ConsSet(N)$, and the *candidate set* of $N$, denoted $CandSet(N)$, are defined recursively:

- If $N$ is a leaf, its consistency set contains all permutations of $\mathcal{S}(N)$:

$$ConsSet(N) = \pi(\mathcal{S}(N)).$$

- If $N$ is an AND-node, and its child nodes are $N_1, N_2, \ldots N_k$, we define the consistency set of $N$ as the set of all possible orderings of $\mathcal{S}(N)$ consistent with candidates of $N_1, N_2, \ldots N_k$:

$$ConsSet(N) = \left\{ \vec{O}_N \in \pi(\mathcal{S}(N)) \;\middle|\; \forall i \; (1 \leq i \leq k), \; \exists \vec{O}_i \in CandSet(N_i) : Cons(\vec{O}_i, \vec{O}_N) \right\}.$$

- If $N$ is an OR-node, and its child node corresponding to every binder $A \in \mathcal{S}(N)$ is $N_A$, then the consistency set of $N$ is obtained by adding binders as the first elements to the candidates of the children:

$$ConsSet(N) = \left\{ A \| \vec{O}_A \ \Big| \ A \in \mathcal{S}(N), \ \vec{O}_A \in CandSet(N_A) \right\}.$$

- A candidate set of $N$ is any set of orderings produced by removing MC-contradicting and MC-equivalent orderings out of the consistency set of $N$, while keeping at least one representative for each group of MC-equivalent orderings:

$$CandSet(N) \subseteq ConsSet(N),$$
$$\vec{O}_N \in (ConsSet(N) \setminus CandSet(N)) \ \Rightarrow \ \mathrm{MCC}_{N, \mathcal{S}_0}(\vec{O}_N) \ \vee$$
$$\left[ \exists \vec{O}'_N \in CandSet(N) : \ \mathrm{MCE}_{N, \mathcal{S}_0}(\vec{O}_N, \vec{O}'_N) \right].$$

(In other words, if some ordering is rejected, it is either MC-contradicting, or MC-equivalent to some other ordering, which is not rejected.)

There are two kinds of orderings which can be removed from $ConsSet(N)$ while retaining its validity: *MC-contradicting* and *MC-equivalent* orderings. Removal of an MC-contradicting ordering cannot change the number of min-consistent orderings in the set; if we remove an MC-equivalent ordering, then even if it is min-consistent, some other min-consistent ordering is retained in the set. If there exists a min-consistent ordering of the set of the node, then its candidate set must contain a min-consistent ordering, and therefore the candidate set is valid.

Note that when our algorithm treats an OR-node, the binder of each child is always placed as the first subgoal of the produced ordering of this node. On higher levels the inner order of subgoals in the ordering does not change (consistency is preserved). Therefore, our algorithm can only produce binder-consistent orderings. This explains the choice of the names *"binder"* and *"binding set"*: the subgoals of $\mathcal{B}(N)$ bind some common variables of $\mathcal{S}(N)$, since they stand to the left of them in any global ordering that our algorithm produces. In particular, if $\mathcal{S}(N)$ is independent under $\mathcal{B}(N)$, then the subgoals of $\mathcal{B}(N)$ bind all the shared free variables of $\mathcal{S}(N)$.

To implement the **DPart** function, we can use the Union-Find data structure (Cormen, Leiserson, & Rivest, 1991, Chapter 22), where subgoals are elements, and indivisible sets are groups. In the beginning, every subgoal constitutes a group by itself. Whenever we discover that two subgoals share a free variable not bound by subgoals of the binding set, we unite their groups into one. To complete the procedure, we need a way to determine which variables are bound by the given binding set. Section 7.1 contains a discussion of this problem and proposes some practical solutions. Finally, we collect all the indivisible subgoals into a separate group. These operations can be implemented in $O(n\alpha(n, n))$ amortized time, where $\alpha(n, n)$ is the inverse Ackermann function, which can be considered $O(1)$ for all values of $n$ that can appear in realistic logic programs. Thus, the whole process of finding the divisibility partition of $n$ subgoals can be performed in $O(n)$ average time.

The formal listing of the ordering algorithm discussed above is shown in Figure 10. The algorithm does not specify explicitly how candidate sets are created from consistency sets. To complete this algorithm, we must provide the three filtering procedures

**Algorithm 5**

**Order**($\mathcal{S}_0$)
    $RootCandSet \leftarrow$ **CandidateSet**($\mathcal{S}_0, \emptyset$)
    Return the cheapest member of $RootCandSet$

**CandidateSet**($\mathcal{S}, \mathcal{B}$)
    case ($\mathcal{S}$ under $\mathcal{B}$)
    <u>independent</u>:
        let $ConsSet_N \leftarrow \pi(\mathcal{S})$
        let $CandSet_N \leftarrow$ **ValidLeafFilter**($ConsSet_N$)
    <u>divisible</u>:
        let $\{\mathcal{S}_1, \mathcal{S}_2, \ldots \mathcal{S}_k\} \leftarrow$ **DPart**($\mathcal{S}, \mathcal{B}$)
        loop for $i = 1$ to $k$
            let $\mathcal{C}_i \leftarrow$ **CandidateSet**($\mathcal{S}_i, \mathcal{B}$)
        let $ConsSet_N \leftarrow \left\{ \vec{O}_N \in \pi(\mathcal{S}(N)) \mid \forall i = 1 \ldots k,\ \exists \vec{O}_i \in \mathcal{C}_i : Cons(\vec{O}_i, \vec{O}_N) \right\}$
        let $CandSet_N \leftarrow$ **ValidANDFilter**($ConsSet_N, \{\mathcal{S}_1, \ldots \mathcal{S}_k\}, \{\mathcal{C}_1, \ldots \mathcal{C}_k\}$)
    <u>indivisible</u>:
        loop for $A \in \mathcal{S}$
            let $\mathcal{C}(A) \leftarrow$ **CandidateSet**($\mathcal{S} \setminus \{A\}, \mathcal{B} \cup \{A\}$)
            let $\mathcal{C}'(A) \leftarrow \left\{ A \| \vec{O}_A \mid \vec{O}_A \in \mathcal{C}(A) \right\}$
        let $ConsSet_N \leftarrow \bigcup_{A \in \mathcal{S}} \mathcal{C}'(A)$
        let $CandSet_N \leftarrow$ **ValidORFilter**($ConsSet_N$)
    Return $CandSet_N$

Figure 10: The skeleton of the DAC ordering algorithm. For each type of node in a divisibility tree, a consistency set is created and refined through validity filters. The produced candidate set of the root is valid; hence, its cheapest member is a minimal ordering of the given set.

– **ValidLeafFilter**, **ValidANDFilter** and **ValidORFilter**. Trivially, we can define them all as null filters that return the sets they receive unchanged. In this case the candidate set of every node will contain all the permutations of its subgoals, and will surely be valid. This will, however, greatly increase the ordering time. Our intention is to reduce the sizes of candidate sets as far as possible, while keeping them valid.

In the following two subsections we discuss the filtering procedures. Section 4.4 discusses detection of MC-contradicting orderings, and Section 4.5 discusses detection of MC-equivalent orderings. Finally, in Section 4.6 we present the complete ordering algorithm, incorporating the filters into the skeleton of Algorithm 5.

## 4.4 Detection of MC-Contradicting Orderings

In this subsection we show sufficient conditions for an ordering to be MC-contradicting. Such orderings can be safely discarded, leaving the set of orderings valid, but reducing its size. The subsection is divided into three parts, one for each type of node in a divisibility tree.

### 4.4.1 DETECTION OF MC-CONTRADICTING ORDERINGS IN LEAVES

The following lemma shows that subgoals in a min-consistent ordering of a leaf node must be sorted by $cn$.

**Lemma 4**
Let $\mathcal{S}_0$ be a set of subgoals, $N$ be a leaf in the divisibility tree of $\mathcal{S}_0$. Let $\vec{O}_N$ be an ordering of $\mathcal{S}(N)$. If the subgoals of $\vec{O}_N$ are not sorted by $cn$ under $\mathcal{B}(N)$, then $\vec{O}_N$ is MC-contradicting.

**Proof:** Let $\vec{O}_S$ be any ordering of $\mathcal{S}_0$, binder-consistent with $\vec{O}_N$. We show that $\vec{O}_S$ cannot be a minimal ordering of $\mathcal{S}_0$, thus $\vec{O}_N$ is not min-consistent.

$\vec{O}_N$ is not sorted by $cn$, i.e., it contains an adjacent cn-inverted pair of subgoals $\langle A_1, A_2 \rangle$. (Recall that a pair is cn-inverted if the first element has a larger $cn$ value than the second one – Section 3.2.3). Since $\vec{O}_S$ is consistent with $\vec{O}_N$, we can write $\vec{O}_S = \vec{X} \| A_1 \| \vec{Y} \| A_2 \| \vec{Z}$, where $\vec{X}$, $\vec{Y}$ and $\vec{Z}$ are (possibly empty) sequences of subgoals. Since $\vec{O}_S$ is binder-consistent with $\vec{O}_N$, $\mathcal{B}(N) \subseteq \vec{X}$.

If $\vec{Y}$ is empty, then $A_1$ and $A_2$ are adjacent in $\vec{O}_S$. Since $\mathcal{B}(N) \subseteq \vec{X}$, $A_1$ and $A_2$ are independent under $\vec{X}$. Therefore, the cost of the whole ordered sequence can be reduced by transposing $A_1$ and $A_2$, according to Lemma 2 (they are adjacent, independent and cn-inverted).

If $\vec{Y}$ is not empty, then no subgoal of $\vec{Y}$ belongs to $\mathcal{S}(N)$, since otherwise it would appear in $\vec{O}_N$ between $A_1$ and $A_2$. By Corollary 2, $\vec{Y}$ is mutually independent of both $A_1$ and $A_2$ under $\vec{X}$.

- If $cn(\vec{Y})|_{\vec{X}} < cn(A_1)|_{\vec{X}}$ then, by Lemma 2, a transposition of $\vec{Y}$ with $A_1$ produces an ordering with lower cost.

- Otherwise, $cn(\vec{Y})|_{\vec{X}} \geq cn(A_1)|_{\vec{X}}$. Since the pair $\langle A_1, A_2 \rangle$ is cn-inverted, $cn(A_1)|_{\vec{X}} > cn(A_2)|_{\vec{X}}$. Hence, $cn(\vec{Y})|_{\vec{X}} > cn(A_2)|_{\vec{X}}$, and transposition of $\vec{Y}$ with $A_2$ reduces the cost, by Lemma 2.

In either case, there is a way to reduce the cost of $\vec{O}_S$. Therefore, $\vec{O}_S$ cannot be minimal, and $\vec{O}_N$ is MC-contradicting. $\qquad\square$

### 4.4.2 DETECTION OF MC-CONTRADICTING ORDERINGS IN AND-NODES

Every member of the consistency set of an AND-node is consistent with some combination of candidates of its child nodes. If there are $k$ child nodes, and for each child $N_i$ the sizes of subgoal and candidate sets are $|\mathcal{S}(N_i)| = n_i$ and $|CandSet(N_i)| = c_i$, then the total number of possible consistent orderings is $c_1 \cdot c_2 \cdot \ldots c_k \cdot \frac{(n_1 + n_2 + \ldots + n_k)!}{n_1! \cdot n_2! \ldots n_k!}$. Fortunately, most of these orderings are MC-contradicting and can be discarded from the candidate set. The following lemma states that it is forbidden to insert other subgoals between two cn-inverted sub-sequences. If such insertion takes place, the ordering is MC-contradicting and can be safely discarded.

**Lemma 5**
*Let $\mathcal{S}_0$ be a set of subgoals, $N$ a node in the divisibility tree of $\mathcal{S}_0$, and $\vec{O}_S$ an ordering of $\mathcal{S}_0$, binder-consistent with an ordering $\vec{O}_N$ of $\mathcal{S}(N)$.*

*If $\vec{O}_N$ contains an adjacent cn-inverted pair of sub-sequences $\langle \vec{A}_1, \vec{A}_2 \rangle$, $\vec{A}_1$ and $\vec{A}_2$ appear in $\vec{O}_S$ not mixed with other subgoals, and $\vec{A}_1$ and $\vec{A}_2$ are not adjacent in $\vec{O}_S$, then $\vec{O}_S$ is not minimal.*

**Proof:** Let $\vec{O}_S$ be such an ordering of $\mathcal{S}_0$, binder-consistent with $\vec{O}_N$:

$$\vec{O}_S = \vec{X} \| \vec{A}_1 \| \vec{Y} \| \vec{A}_2 \| \vec{Z},$$

where $\vec{Y}$ is not empty. No subgoal of $\vec{Y}$ belongs to $\mathcal{S}(N)$, since otherwise it would stand in $\vec{O}_N$ between $\vec{A}_1$ and $\vec{A}_2$. $\vec{O}_S$ is binder-consistent with $\vec{O}_N$; therefore, $\mathcal{B}(N) \subseteq \vec{X}$. By Corollary 2, $\vec{Y}$ must be mutually independent of both $\vec{A}_1$ and $\vec{A}_2$ under $\vec{X}$, and by Lemma 2 a transposition of $\vec{Y}$ with either $\vec{A}_1$ or $\vec{A}_2$ reduces the cost – exactly as in the proof of Lemma 4. $\square$

If a pair of adjacent subgoals $\langle A_i, A_{i+1} \rangle$ is cn-inverted, then by the previous lemma any attempt to insert subgoals inside it results in a non-minimal global ordering. Thereupon we may join $A_i$ and $A_{i+1}$ into a *block* $A_{i,i+1}$, which can further participate in a larger block. The formal recursive definition of a block follows. For convenience, we consider separate subgoals to be blocks of length 1.

**Definition:**

1. A sub-sequence $\vec{A}$ of an ordered sequence of subgoals is a *block* if it is either a single subgoal, or $\vec{A} = \vec{A}_1 \| \vec{A}_2$, where $\langle \vec{A}_1, \vec{A}_2 \rangle$ is a cn-inverted pair of blocks.

2. A block is *maximal* (*max-block*) if it is not a sub-sequence of a larger block.

3. Let $N$ be a node in a divisibility tree, $M$ be some descendant of $N$, $\vec{O}_N \in \pi(\mathcal{S}(N))$ and $\vec{O}_M \in \pi(\mathcal{S}(M))$ be two consistent orderings of these nodes. A block $\vec{A}$ of $\vec{O}_M$ is *violated* in $\vec{O}_N$ if there are two adjacent subgoals in $\vec{A}$ that are not adjacent in $\vec{O}_N$ (in other words, alien subgoals are inserted between the subgoals of the block).

4. Let $N$ be a node, $M$ be its descendant, $\vec{O}_N \in \pi(\mathcal{S}(N))$ and $\vec{O}_M \in \pi(\mathcal{S}(M))$ be two consistent orderings of these nodes. $\vec{O}_M$ is called the *projection* of $\vec{O}_N$ on $M$. We shall usually speak about projection of an ordering on a child node.

The concept of max-block is similar to the *maximal indivisible block* introduced by Simon and Kadane (1975) in the context of satisficing search. The following corollary presents the result of Lemma 5 in a more convenient way.

**Corollary 3** *Let $N$ be a node in a divisibility tree, $M$ be one of its children, $\vec{O}_N$ be an ordering of $N$, and $\vec{O}_M$ be the projection of $\vec{O}_N$ on $M$. If $\vec{O}_M$ contains a block that is violated in $\vec{O}_N$, then $\vec{O}_N$ is MC-contradicting.*

**Proof:** Let $\vec{A}$ be the smallest block of $\vec{O}_M$ violated in $\vec{O}_N$. According to the definition of a block, $\vec{A} = \vec{A}_1 \| \vec{A}_2$, where $\vec{A}_1$ and $\vec{A}_2$ are not violated in $\vec{O}_N$, and the pair $\langle \vec{A}_1, \vec{A}_2 \rangle$ is cn-inverted. Let $\vec{O}_S$ be any ordering of the root node binder-consistent with $\vec{O}_N$. $\vec{O}_S$ violates $\vec{A}$, since $\vec{O}_N$ violates $\vec{A}$. To show that $\vec{O}_N$ is MC-contradicting, we must prove that $\vec{O}_S$ is not minimal.

- If $\vec{A}_1$ and $\vec{A}_2$ are not violated in $\vec{O}_S$, then they are not adjacent in $\vec{O}_S$, and $\vec{O}_S$ is not minimal, by Lemma 5.

- Otherwise, $\vec{A}_1$ or $\vec{A}_2$ is violated in $\vec{O}_S$. Without loss of generality, let it be $\vec{A}_1$. Let $\vec{A}'$ be the smallest sub-block of $\vec{A}_1$ violated in $\vec{O}_S$. According to the definition of a block, $\vec{A}' = \vec{A}'_1 \| \vec{A}'_2$, where the pair $\langle \vec{A}'_1, \vec{A}'_2 \rangle$ is cn-inverted, $\vec{A}_1$ and $\vec{A}_2$ are not violated and not adjacent in $\vec{O}_S$. By Lemma 5, $\vec{O}_S$ is not minimal. □

For example, if control values of subgoals are as shown in Figure 9, then $\langle a1(X), a2(X) \rangle$ is a block, since $cn(a1(X))|_{\emptyset} = \frac{2-1}{2} = \frac{1}{2}$, $cn(a2(X))|_{\{a1(X)\}} = \frac{2-1}{3} = \frac{1}{3}$. As one can see from the figure, insertion of $b$ or $d$ inside this block results in a non-minimal ordering.

As was already noted above, the consistency set of an AND-node can be large. In many of its orderings, however, blocks of projections are violated, and we can discard these orderings as MC-contradicting. In the remaining orderings, no block of a projection is violated, and each such ordering can be represented as a sequence of max-blocks of the projections. In each projection, its max-blocks stand in $cn$-ascending order (otherwise, there is an adjacent cn-inverted pair of blocks, and a larger block can be formed, which contradicts their maximality). As the following lemma states, in the parent AND-node these blocks must also be ordered by their $cn$ values; otherwise, the ordering is MC-contradicting.

**Lemma 6** *If an ordering of an AND-node contains an adjacent cn-inverted pair of max-blocks of its projections on the children, then this ordering is MC-contradicting.*

**Proof:** If these blocks are violated in the binder-consistent global ordering, the global ordering is not minimal by Corollary 3. If the blocks are not violated, the proof is similar to the proof of Lemma 4. □

The two sufficient conditions for detection of MC-contradicting orderings expressed in Corollary 3 and Lemma 6 allow us to reduce the size of the candidate set significantly. Assume, for example, that the set of our current node $N$ is split into two mutually independent subsets whose candidates are $\langle a_1, a_2 \rangle$ and $\langle b_1, b_2 \rangle$ (one candidate for each child). There are six possible orderings of $\mathcal{S}(N)$, all shown in Figure 11. Assume that both $\langle a_1, a_2 \rangle$ and $\langle b_1, b_2 \rangle$ are blocks, and $cn(\langle a_1, a_2 \rangle)|_{\mathcal{B}(N)} < cn(\langle b_1, b_2 \rangle)|_{\mathcal{B}(N)}$. Out of six consistent orderings, four (2–5) can be rejected due to block violation, and one of the remaining two (number 6) puts the blocks in the wrong order. So, only one ordering (number 1) can be left in the candidate set of $N$. Even if neither $\langle a_1, a_2 \rangle$ nor $\langle b_1, b_2 \rangle$ are blocks, Lemma 6 dictates a unique interleaving of their elements (max-blocks), assuming that $cn(a_1)|_{\mathcal{B}(N)} \neq cn(a_2)|_{\mathcal{B}(N) \cup \{a_1\}} \neq cn(b_1)|_{\mathcal{B}(N)} \neq cn(b_2)|_{\mathcal{B}(N) \cup \{b_1\}}$.

### 4.4.3 DETECTION OF MC-CONTRADICTING ORDERINGS IN OR-NODES

The following lemma states that if a block has a cheaper permutation, then the ordering is MC-contradicting (and can be discarded from the candidate set).

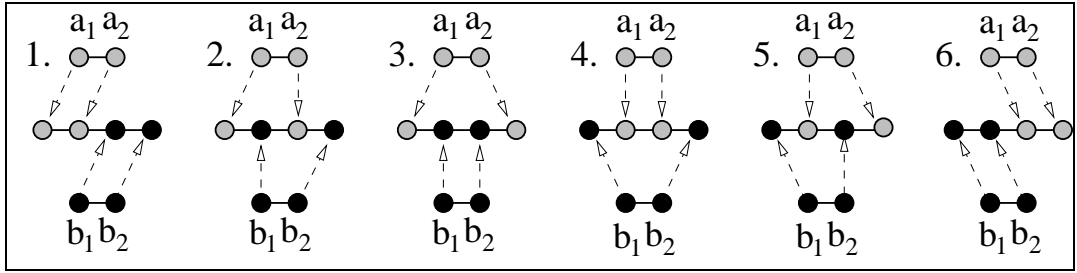

Figure 11: The possible ways to combine $\langle a_1, a_2 \rangle$ and $\langle b_1, b_2 \rangle$

**Lemma 7** *Let $N$ be a node in the divisibility tree of $\vec{S}_0$, $\vec{O}_N \in \pi(\vec{S}(N))$. Let $\vec{A}$ be a leading block of $\vec{O}_N$: $\vec{O}_N = \vec{A} \| \vec{R}$. If there is a permutation of $\vec{A}$, $\vec{A}'$, such that $\overline{cost}(\vec{A}')|_{\mathcal{B}(N)} < \overline{cost}(\vec{A})|_{\mathcal{B}(N)}$, then $\vec{O}_N$ is MC-contradicting.*

**Proof:** Let $\vec{O}_S \in \pi(\vec{S}_0)$ be binder-consistent with $\vec{O}_N$. If $\vec{A}$ is violated in $\vec{O}_S$, $\vec{O}_S$ cannot be minimal (Corollary 3). Otherwise, $\vec{A}$ occupies a continuous segment in $\vec{O}_S$, and its replacement by a cheaper permutation reduces the cost of the global ordering (Lemma 1). Thus, $\vec{O}_S$ cannot be minimal. □

This check should be done only for leading blocks of OR-nodes:

- Every ordering of a leaf node that has not been rejected due to Lemma 4 must be sorted by $cn$. Consequently, it contains no cn-inverted adjacent pair of subgoals, and no block of size $\geq 2$ can be formed.

- Every ordering of an AND-node that has not been rejected due to Corollary 3 or Lemma 6 must have its blocks unbroken and in cn-ascending order. Consequently, new blocks cannot be formed here either.

- In OR-nodes, new blocks can be formed when we add a binder as the first element of an ordering, if the $cn$ value of the binder is greater than that of the subsequent block. All new blocks start from the binder, and we must perform the permutation test only on the leading max-block of an ordering.

### 4.5 Detection of MC-Equivalent Orderings

In the previous subsection we presented sufficient conditions for detecting MC-contradicting orderings. In this subsection we specify sufficient conditions for identifying MC-equivalent orderings. Recall that two orderings of a node are MC-equivalent if minimal consistency of one implies minimal consistency of the other. Finding such sufficient conditions will allow us to eliminate orderings without loss of validity of the candidate set. We start with defining a specialization of the MC-equivalence relation: *blockwise equivalence*. We then show that orderings whose max-blocks are sorted by $cn$ are blockwise-equivalent, and therefore MC-equivalent.

**Definition:** Let $\mathcal{S}_0$ be a set of subgoals and $N$ be a node in the divisibility tree of $\mathcal{S}_0$. Let $\vec{O}_1$ and $\vec{O}_2$ be two orderings of $\mathcal{S}(N)$ with an equal number of max-blocks. Let $\vec{O}_S$ be an ordering of $\mathcal{S}_0$, binder-consistent with $\vec{O}_1$, where blocks of $\vec{O}_1$ are not violated.

$\vec{O}_S|_{\vec{O}_1}^{\vec{O}_2}$ is the ordering obtained by replacing in $\vec{O}_S$ every max-block of $\vec{O}_1$ with a max-block of $\vec{O}_2$, while preserving the order of max-blocks (the $i$-th max-block of $\vec{O}_1$ is replaced by the $i$-th max-block of $\vec{O}_2$).

$\vec{O}_1$ and $\vec{O}_2$ are *blockwise-equivalent* if the following condition holds: $\vec{O}_1$ is min-consistent with $\vec{O}_S$ iff $\vec{O}_2$ is min-consistent with $\vec{O}_S|_{\vec{O}_1}^{\vec{O}_2}$.

As can be easily seen, if two orderings are blockwise-equivalent, then they are MC-equivalent. Now we show that a transposition of adjacent, mutually independent cn-equal max-blocks in an ordering of a node produces a blockwise-equivalent ordering. The proof of the following lemma is found in Appendix A.

**Lemma 8**
*Let $\mathcal{S}_0$ be a set of subgoals, $N$ be a node in the divisibility tree of $\mathcal{S}_0$, $\vec{O}_N = \vec{Q} \| \vec{A}_1 \| \vec{A}_2 \| \vec{R}$ be an ordering of $\mathcal{S}(N)$, where $\vec{A}_1$ and $\vec{A}_2$ are max-blocks, mutually independent and cn-equal under the bindings of $\mathcal{B}(N) \cup \vec{Q}$. Then $\vec{O}_N$ is blockwise-equivalent with $\vec{O}'_N = \vec{Q} \| \vec{A}_2 \| \vec{A}_1 \| \vec{R}$.*

**Corollary 4** *All sorted by cn orderings of a leaf node are blockwise-equivalent.*

For example, if $\mathcal{S}(N) = \{A, B, C, D\}$, $cn(A)|_{\mathcal{B}(N)} = 0.1$, $cn(B)|_{\mathcal{B}(N)} = cn(C)|_{\mathcal{B}(N)} = 0.3$, $cn(D)|_{\mathcal{B}(N)} = 0.5$, then the orderings $\langle A, B, C, D \rangle$ and $\langle A, C, B, D \rangle$ are blockwise-equivalent, and we can remove from the candidate set any one of them (but not both).

**Corollary 5** *All orderings of an AND-node, where blocks of projections are not violated and adjacent max-blocks from different children projections are cn-ordered, are blockwise-equivalent.*

For example, if the candidates of the children are $\vec{A} \| \vec{B}$ and $\vec{C} \| \vec{D}$, where $\vec{A}, \vec{B}, \vec{C}, \vec{D}$ are max-blocks, $cn(\vec{A})|_{\mathcal{B}(N)} = 0.1$, $cn(\vec{B})|_{\mathcal{B}(N) \cup \vec{A}} = cn(\vec{C})|_{\mathcal{B}(N)} = 0.3$ and $cn(\vec{D})|_{\mathcal{B}(N) \cup \vec{C}} = 0.5$, then the orderings $\vec{A} \| \vec{B} \| \vec{C} \| \vec{D}$ and $\vec{A} \| \vec{C} \| \vec{B} \| \vec{D}$ are blockwise-equivalent, and we can remove from the candidate set any one of them (but not both).

To prove both Corollaries 4 and 5, we note that in each case one of the mentioned orderings can be obtained from the other by a finite number of transpositions of adjacent, mutually independent and cn-equal max-blocks. According to Lemma 8, each such transposition yields a blockwise-equivalent ordering. It is easy to show that blockwise equivalence is transitive.

The following corollary states that subgoals within a block can be permuted, provided that the cost of the block is not changed.

**Corollary 6** *All orderings of a node, identical up to cost-preserving permutations of subgoals inside blocks, are blockwise-equivalent.*

The proof of the corollary follows immediately from Lemma 1. For example, if the set is $\{a(X), b(X)\}$, and the control values are as in the first counter-example of Proposition 1, i.e. $cn(a(X)|_\emptyset) = cn(b(X)|_\emptyset) = \frac{1}{2}$, and $cn(a(X)|_{\{b(X)\}}) = cn(b(X)|_{\{a(X)\}}) = 0$, then in both possible orderings, $\langle a(X), b(X) \rangle$ and $\langle b(X), a(X) \rangle$, the two subgoals are united into a block, and these blocks have equal cost. In any global ordering containing the block $\langle a(X), b(X) \rangle$, we can replace this block with $\langle b(X), a(X) \rangle$ without changing the total cost. Therefore $\langle a(X), b(X) \rangle$ is blockwise-equivalent to $\langle b(X), a(X) \rangle$.

| Node | Set | MC-contradicting | blockwise-equivalent |
|------|-----|------------------|----------------------|
| Leaf | Independent | Subgoals not sorted by $cn$ <br> — *Lemma 4* | Subgoals sorted by $cn$ <br> — *Corollary 4* |
| AND | Divisible | Contains violated blocks <br> — *Corollary 3* <br> Max-blocks not sorted by $cn$ <br> — *Lemma 6* | Max-blocks not violated, <br> sorted by $cn$ <br> — *Corollary 5* |
| OR | Indivisible | The leading max-block has <br> a cheaper permutation <br> — *Lemma 7* | Cost-preserving permutations <br> of blocks <br> — *Corollary 6* |

Table 1: Summary of sufficient conditions for detection of MC-contradicting and blockwise-equivalent orderings.

The sufficient condition expressed in Corollary 6 should be checked only in OR-nodes, since in leaves and AND-nodes no new blocks are created, as was argued in Section 4.4.3.

## 4.6 The Revised Ordering Algorithm

In the two preceding subsections we saw several sufficient conditions of MC-contradiction and MC-equivalence, summarized in Table 1. These results permit us to close the gaps in Algorithm 5 by providing the necessary validity filters. Each filter tests the sufficient conditions of MC-contradiction and MC-equivalence on every ordering in the consistency set. If some of these sufficient conditions hold, the ordering is rejected. The formal listing of these procedures is shown in Figure 12.

While the generate-and-test approach described above served us well for methodological purposes, it is obviously not practical because of its computational limitations. For example, for an independent set of size $n$, the algorithm creates $n!$ orderings, then rejects $n! - 1$ and keeps only one. This process takes $O(n! \cdot n)$ time and produces an ordering which is sorted by $cn$. The same result could be obtained in just $O(n \log n)$ time, by a single sorting. So, instead of uncontrolled creation of orderings and selective rejection, we want to perform a selective creation of orderings. In other words, we want to revise our algorithm to deal directly with candidate sets, instead of generating large consistency sets. The revised algorithm produces the candidate set of a node $N$ as follows:

- If $N$ is a leaf, the subgoals of $\mathcal{S}(N)$ are sorted by $cn$ under the bindings of $\mathcal{B}(N)$, and the produced ordering is the sole candidate of $N$.

- If $N$ is an AND-node, then for each combination of its children's candidates a candidate of $N$ is created, where the max-blocks of the children's candidates are ordered

**ValidLeafFilter**($ConsSet_N$)

    let $CandSet_N \leftarrow \emptyset$

    loop for $\vec{O}_N \in ConsSet_N$

        if $\vec{O}_N$ is sorted by $cn$

        and there is no $\vec{O}'_N \in CandSet_N$ which is sorted by $cn$

        then $CandSet_N \leftarrow CandSet_N \cup \{\vec{O}_N\}$

    Return $CandSet_N$

**ValidANDFilter**($ConsSet_N, \{\mathcal{S}_1, \dots \mathcal{S}_k\}, \{\mathcal{C}_1, \dots \mathcal{C}_k\}$)

    let $CandSet_N \leftarrow \emptyset$

    loop for $\vec{O}_N \in ConsSet_N$

        loop for $i = 1$ to $k$

            let $\vec{O}_i$ be the projection of $\vec{O}_N$ on $\mathcal{S}_i$

        if $\forall i \ \vec{O}_i \in \mathcal{C}_i$

        and max-blocks of $\vec{O}_i$-s are not violated in $\vec{O}_N$,

        and max-blocks of $\vec{O}_i$-s are ordered by $cn$ in $\vec{O}_N$,

        and there is no $\vec{O}'_N \in CandSet_N$ consistent with all $\vec{O}_i$-s,

        then $CandSet_N \leftarrow CandSet_N \cup \{\vec{O}_N\}$

    Return $CandSet_N$

**ValidORFilter**($ConsSet_N$)

    let $CandSet_N \leftarrow \emptyset$

    loop for $\vec{O}_N \in ConsSet_N$

        if $\vec{O}_N$ does not start with a block having a cheaper permutation,

        and there is no $\vec{O}'_N \in CandSet_N$, identical to $\vec{O}_N$ up to

            cost-preserving permutations in blocks,

        then $CandSet_N \leftarrow CandSet_N \cup \{\vec{O}_N\}$

    Return $CandSet_N$

Figure 12: The three filter procedures that convert a consistency set into a candidate set. Together with Algorithm 5, they form a complete ordering algorithm. The efficiency of the algorithm can be improved, as we shall see in Algorithm 6.

by $cn$. The candidate is produced by merging: moving in parallel on the candidates of the children and extracting max-blocks that are minimal by $cn$.

- If $N$ is an OR-node, then for each candidate of its child an ordering of $N$ is created by adding the binder to the left end of the child candidate. If this results in creation of a block that has a cheaper permutation, the ordering is rejected; otherwise, it is added to the candidate set. It suffices to check only the leading max-block.

Note that the revised algorithm does not include a test for cost-preserving permutations of blocks in different orderings (expressed in Corollary 6), because of the high expense of such a test.

The revised algorithm described above contains manipulations of blocks. For this purpose, we need an easy and efficient way to detect blocks in orderings. Since we do not permit block violation (by Corollary 3), we can unite all the subgoals of a max-block into one entity, and treat it as an ordinary subgoal. The procedure of joining subgoals into blocks is called *folding*, and the resulting sequence of max-blocks – a *folded sequence*. After subgoals are folded into a block, there is no need to unfold this block back to separate subgoals: on upper levels of the tree, these subgoals will again be joined into a block, unless the block is violated. The unfolding operation is carried out only once before returning the cheapest ordering of the set (of the root node). The candidate sets of the nodes are now defined as sets of *folded* orderings.

As was already stated, new blocks can only be created in the candidates of OR-nodes, when the binder is added as the first element of the ordering, if the *cn* value of the binder is greater than the *cn* value of the first max-block of the child projection. Therefore, in the revised algorithm we only build new blocks that start from the binder: the max-blocks in the rest of the ordering remain from the child's candidate. First we try to make a block out of the binder and the first max-block of the child's candidate. If they are cn-ordered, we stop the folding. If they are cn-inverted, we unite them into a larger block, and try to unite it with the second max-block of the child's candidate, and so on. The produced folded ordering contains only maximal blocks: the first block is maximal, since we could not expand it further to the right, and the other blocks are maximal, since they were maximal in the child's candidate.

Lemma 7 states that an ordering whose leading max-block has a cheaper permutation is MC-contradicting. One way to detect such a block is to exhaustively test all its permutations, computing and comparing their costs. This procedure is very expensive. Instead, in our revised algorithm we employ the adjacency restriction test (Equation 8). The test is applied to every pair of adjacent subgoals of a block, and if some adjacent pair has a cheaper transposition, then the whole block has a cheaper permutation, by Lemma 1. Since blocks are created by concatenation of smaller blocks, it suffices to test the adjacency restriction only at the points where blocks are joined (for other adjacent pairs of subgoals, the tests were performed on the lower levels, when smaller blocks were formed). The adjacency restriction test does not guarantee detection of all not-cheapest permutations (as was shown in Example 3), but it detects such blocks in many cases, and works in linear time.

The final version of the DAC subgoal ordering algorithm is presented in Figure 13. The complete correctness proof of Algorithm 6 is found in Appendix B.

## 4.7 Sample Run and Comparison of Ordering Algorithms

We illustrate the work of the DAC algorithm, using the subgoal set shown in Figure 8, $\mathcal{S}_0 = \{a, b, c(X), d(X), e(X)\}$. After proving $c(X)$, $d(X)$ or $e(X)$, we can assume that $X$ is bound. Let the control values for the subgoals be as shown in Table 2. The column $c(free)$ contains control values for the subgoal $c(X)$ when $X$ is not yet bound by the preceding subgoals (i.e., the binding set does not contain $d(X)$ or $e(X)$). The column $c(bound)$

---

**Algorithm 6** : *The Divide-and-Conquer Algorithm*

**Order**($\mathcal{S}_0$)
    let *RootCandSet* $\leftarrow$ **CandidateSet**($\mathcal{S}_0, \emptyset$)
    Return **Unfold**(the cheapest element of *RootCandSet*)

**CandidateSet**($\mathcal{S}, \mathcal{B}$)
    let $\{\mathcal{S}_1, \mathcal{S}_2, \ldots \mathcal{S}_k\} \leftarrow$ **DPart**($\mathcal{S}, \mathcal{B}$)
    case
        • $k = 1$, *shared-vars*($\mathcal{S}_1$) $= \emptyset$ ($\mathcal{S}$ is <u>independent</u> under $\mathcal{B}$):
            Return $\{$**Sort-by-cn**($\mathcal{S}, \mathcal{B}$)$\}$
        • $k = 1$, *shared-vars*($\mathcal{S}_1$) $\neq \emptyset$ ($\mathcal{S}$ is <u>indivisible</u> under $\mathcal{B}$):
            loop for $A \in \mathcal{S}$
                let $\mathcal{C}(A) \leftarrow$ **CandidateSet**($\mathcal{S} \setminus \{A\}, \mathcal{B} \cup \{A\}$)
                let $\mathcal{C}'(A) \leftarrow \left\{ \textbf{Fold}(A \| \vec{O}_A, \mathcal{B}) \mid \vec{O}_A \in \mathcal{C}(A) \right\}$
            Return $\bigcup_{A \in \mathcal{S}} \mathcal{C}'(A)$
        • $k > 1$ ($\mathcal{S}$ is <u>divisible</u> under $\mathcal{B}$):
            loop for $i = 1$ to $k$
                let $\mathcal{C}_i \leftarrow$ **CandidateSet**($\mathcal{S}_i, \mathcal{B}$)
            Return $\left\{ \textbf{Merge}(\{\vec{O}_1, \vec{O}_2, \ldots \vec{O}_k\}, \mathcal{B}) \mid \vec{O}_1 \in \mathcal{C}_1, \vec{O}_2 \in \mathcal{C}_2, \ldots \vec{O}_k \in \mathcal{C}_k \right\}$

**Merge**($\{\vec{O}_1, \vec{O}_2, \ldots \vec{O}_k\}, \mathcal{B}$)
    let *min-cn-candidate* $\leftarrow \vec{O}_i$ that minimizes $cn(\text{first-max-block}(\vec{O}_i))|_{\mathcal{B}}$, $1 \leq i \leq k$
    let *min-cn-block* $\leftarrow$ first-max-block(*min-cn-candidate*)
    remove-first-max-block(*min-cn-candidate*)
    Return *min-cn-block*$\|$**Merge**($\{\vec{O}_1, \vec{O}_2, \ldots \vec{O}_k\}, \mathcal{B} \cup$ *min-cn-block*)

**Fold**($\langle A_1, A_2 \ldots A_k \rangle, \mathcal{B}$)
    if $k \leq 1$ or $cn(A_1)|_{\mathcal{B}} \leq cn(A_2)|_{\mathcal{B} \| A_1}$
    then Return $\langle A_1, A_2 \ldots A_k \rangle$
    else
        if the last subgoal of $A_1$ and the first subgoal of $A_2$ satisfy the adjacency restriction
        then
            let $A' \leftarrow$ **block**($A_1, A_2$)
            Return **Fold**($\langle A', A_3 \ldots A_k \rangle, \mathcal{B}$)
        else Return $\emptyset$

---

Figure 13: The revised version of the DAC algorithm. The candidate sets are built selectively, without explicit creation of consistency sets. Candidate sets contain folded orderings, and unfolding is performed only on the returned global ordering. The code of the **Unfold** and **Sort-by-cn** procedures is not listed, due to its straightforwardness. The merging procedure recursively extracts from the given folded orderings max-blocks that are minimal by $cn$. The folding procedure joins two leading blocks into a larger one, as long as they are cn-inverted.

| | a | b | c(free) | c(bound) | d(free) | d(bound) | e(free) | e(bound) |
|---|---|---|---|---|---|---|---|---|
| $c\overline{os}t$ | 10 | 5 | 5 | 5 | 10 | 5 | 20 | 10 |
| $n\overline{s}ols$ | 0.8 | 2 | 2 | 0.5 | 4 | 1 | 0.4 | 0.1 |
| $cn$ | -0.02 | 0.2 | 0.2 | -0.1 | 0.3 | 0 | -0.03 | -0.09 |

Table 2: Control values for the sample runs of the ordering algorithms.

contains cost values of $c(X)$ when $d(X)$ or $e(X)$ have already bound $X$. For example, $c\overline{os}t(c(X))|_{\{a,d(X)\}} = c\overline{os}t(c(bound)) = 5$. The DAC algorithm traverses the divisibility tree of $\mathcal{S}_0$ as follows. (The names of the nodes are as in Figure 8.)

1. The root of the divisibility tree, $n1$, has empty binding set $\mathcal{B}(n1) = \emptyset$, and the associated subgoal set $\mathcal{S}(n1) = \{a, b, c(X), d(X), e(X)\}$. The set $\mathcal{S}(n1)$ is partitioned into two subsets under $\mathcal{B}(n1)$: one independent $- \{a, b\}$, and one indivisible $- \{c(X), d(X), e(X)\}$. These two subsets correspond to two child nodes of the AND-node $n1$: $n2$ and $n3$, both with empty binding sets.

2. $\mathcal{S}(n2)$ is independent under $\mathcal{B}(n2)$. Therefore, $n2$ is a leaf, and its sole candidate ordering is obtained by sorting its subgoals by $cn$ under $\mathcal{B}(n2)$. $cn(a)|_{\emptyset} = -0.02$, $cn(b)|_{\emptyset} = 0.2$, thus $CandSet(n2) = \{\langle a, b \rangle\}$.

3. $\mathcal{S}(n3)$ is indivisible under $\mathcal{B}(n3)$. Therefore, $n3$ is an OR-node, and its three children are created $-$ one for each subgoal of $\mathcal{S}(n3)$ serving as the binder.

   - Binder $c(X)$ yields the child node $n4$ with the associated set $\mathcal{S}(n4) = \{d(X), e(X)\}$ and the binding set $\mathcal{B}(n4) = \{c(X)\}$. $\mathcal{S}(n4)$ is independent under $\mathcal{B}(n4)$. Therefore, $n4$ is a leaf, and its sole candidate is obtained by sorting its subgoals by $cn$:

     $$cn(d(X))|_{\{c(X)\}} = 0, \quad cn(e(X))|_{\{c(X)\}} = -0.09;$$

     thus, the candidate of $n4$ is $\langle e(X), d(X) \rangle$.

   - Binder $d(X)$ yields the child node $n5$ with the associated set $\mathcal{S}(n5) = \{c(X), e(X)\}$ and the binding set $\mathcal{B}(n5) = \{d(X)\}$. $\mathcal{S}(n5)$ is independent under $\mathcal{B}(n5)$, and its sorting by $cn$ produces the candidate $\langle c(X), e(X) \rangle$.

   - Binder $e(X)$ yields the child node $n6$ with the associated set $\mathcal{S}(n6) = \{c(X), d(X)\}$ and the binding set $\mathcal{B}(n6) = \{e(X)\}$. $\mathcal{S}(n6)$ is independent under $\mathcal{B}(n6)$, and its sorting by $cn$ produces the candidate $\langle c(X), d(X) \rangle$.

4. We now add each binder to its corresponding child's candidate and obtain three orderings of the OR-node $n3$: $\langle c(X), e(X), d(X) \rangle$, $\langle d(X), c(X), e(X) \rangle$, $\langle e(X), c(X), d(X) \rangle$.

5. We now perform folding of these orderings and check violations of the adjacency restriction, in order to determine whether a block has a cheaper permutation.

- First, we perform the folding of $\langle c(X), e(X), d(X)\rangle$. The pair $\langle c(X), e(X)\rangle$ is cn-inverted: $cn(c(X))|_\emptyset = 0.2$, $cn(e(X))|_{\{c(X)\}} = -0.09$. We thus unite it into a block. This block does not pass the adjacency restriction test (Equation 8):

$$\begin{aligned} c\overline{os}t(\langle c(X), e(X)\rangle)|_\emptyset &= 5 + 2 \cdot 10 = 25, \\ c\overline{os}t(\langle e(X), c(X)\rangle)|_\emptyset &= 20 + 0.4 \cdot 5 = 22. \end{aligned}$$

Therefore, this ordering is MC-contradicting and can be discarded.

- We perform the folding of $\langle d(X), c(X), e(X)\rangle$. $cn(d(X))|_\emptyset = 0.3$, $cn(c(X))|_{\{d(X)\}} = -0.1$, the pair is cn-inverted, and we unite it into a block. This block does not pass the adjacency restriction test:

$$\begin{aligned} c\overline{os}t(\langle d(X), c(X)\rangle)|_\emptyset &= 10 + 4 \cdot 5 = 30, \\ c\overline{os}t(\langle c(X), d(X)\rangle)|_\emptyset &= 5 + 2 \cdot 5 = 15. \end{aligned}$$

This ordering is rejected too, even before its folding is finished. If we continue the folding process, we shall see that the subgoal $e(X)$ must also be added to this block, since $cn(\langle d(X), c(X)\rangle)|_\emptyset = \frac{4 \times 0.5 - 1}{30} = 0.0333$, and $cn(e(X))|_{\langle d(X), c(X)\rangle} = -0.09$.

- We perform the folding of $\langle e(X), c(X), d(X)\rangle$. $cn(e(X))|_\emptyset = -0.03$, $cn(c(X))|_{\{e(X)\}} = -0.1$, the pair is cn-inverted, and we form a block $ec(X) = \langle e(X), c(X)\rangle$, which passes the adjacency restriction test:

$$\begin{aligned} c\overline{os}t(\langle e(X), c(X)\rangle)|_\emptyset &= 20 + 0.4 \cdot 5 = 22, \\ c\overline{os}t(\langle c(X), e(X)\rangle)|_\emptyset &= 5 + 2 \cdot 10 = 25. \end{aligned}$$

We compute the control values of the new block:

$$\begin{aligned} c\overline{os}t(ec(X))|_\emptyset &= 20 + 0.4 \cdot 5 = 22 \\ n\overline{sol}s(ec(X))|_\emptyset &= 0.4 \cdot 0.5 = 0.2 \\ cn(ec(X))|_\emptyset &= \frac{0.2 - 1}{22} = -0.0363636 \end{aligned}$$

$cn(d(X))|_{\{ec(X)\}} = 0$, thus the pair $\langle ec(X), d(X)\rangle$ is cn-ordered, no more folding is needed, and we add the folded candidate $\langle ec(X), d(X)\rangle$ to the candidate set of $n3$.

6. We now perform merging of the candidate set of $n2$, $\{\langle a, b\rangle\}$, with the candidate set of $n3$, $\{\langle ec(X), d(X)\rangle\}$. In the resulting sequence max-blocks must be sorted by $cn$.

   $cn(a) = -0.02$, $cn(b) = 0.2$, $cn(ec(X))|_\emptyset = -0.0363636$, $cn(d(X))|_{\{ec(X)\}} = 0$.

   The merged ordering, $\langle ec(X), a, d(X), b\rangle$, is added to the candidate set of $n1$.

7. We compare the costs of all candidates of $n1$, and output the cheapest one. In our case, there is only one candidate, $\langle ec(X), a, d(X), b\rangle$. The algorithm returns this candidate unfolded, $\langle e(X), c(X), a, d(X), b\rangle$.

| Cheapest prefix | Extension/Completion | Cost |
|---|---|---|
| $\emptyset$ | $\langle a \rangle$ | 10 |
| | $\langle b \rangle$ | 5 |
| | $\langle c(X) \rangle$ | 5 |
| | $\langle d(X) \rangle$ | 10 |
| | $\langle e(X) \rangle$ | 20 |
| $\langle b \rangle$ | $\langle b, a \rangle$ | the adjacency restriction test fails |
| | $\langle b, c(X) \rangle$ | $5 + 2 \cdot 5 = 15$ |
| | $\langle b, d(X) \rangle$ | $5 + 2 \cdot 10 = 25$ |
| | $\langle b, e(X) \rangle$ | the adjacency restriction test fails |
| $\langle c(X) \rangle$ | $\langle c(X), e(X), a, d(X), b \rangle$ | $5 + 2(10 + 0.1(10 + 0.8(5 + 1 \cdot 5))) = 28.6$ |
| $\langle a \rangle$ | $\langle a, b \rangle$ | $10 + 0.8 \cdot 5 = 14$ |
| | $\langle a, c(X) \rangle$ | $10 + 0.8 \cdot 5 = 14$ |
| | $\langle a, d(X) \rangle$ | $10 + 0.8 \cdot 10 = 18$ |
| | $\langle a, e(X) \rangle$ | the adjacency restriction test fails |
| $\langle d(X) \rangle$ | $\langle d(X), c(X), e(X), a, b \rangle$ | $10 + 4(5 + 0.5(10 + 0.1(10 + 0.8 \cdot 5))) = 52.8$ |
| $\langle a, b \rangle$ | $\langle a, b, c(X) \rangle$ | $14 + 0.8 \cdot 2 \cdot 5 = 22$ |
| | $\langle a, b, d(X) \rangle$ | $14 + 0.8 \cdot 2 \cdot 10 = 30$ |
| | $\langle a, b, e(X) \rangle$ | the adjacency restriction test fails |
| $\langle a, c(X) \rangle$ | $\langle a, c(X), e(X), d(X), b \rangle$ | $14 + 0.8 \cdot 2(10 + 0.1(5 + 1 \cdot 5)) = 31.6$ |
| $\langle b, c(X) \rangle$ | $\langle b, c(X), e(X), a, d(X) \rangle$ | $15 + 2 \cdot 2(10 + 0.1(10 + 0.8 \cdot 5)) = 60.6$ |
| $\langle a, d(X) \rangle$ | $\langle a, d(X), c(X), e(X), b \rangle$ | $18 + 0.8 \cdot 4(5 + 0.5(10 + 0.1 \cdot 5)) = 50.8$ |
| $\langle e(X) \rangle$ | $\langle e(X), c(X), a, d(X), b \rangle$ | $20 + 0.4(5 + 0.5(10 + 0.8(5 + 1 \cdot 5))) = 25.6$ |
| $\langle a, b, c(X) \rangle$ | $\langle a, b, c(X), e(X), d(X) \rangle$ | $22 + 0.8 \cdot 2 \cdot 2(10 + 0.1 \cdot 5) = 55.6$ |
| $\langle b, d(X) \rangle$ | $\langle b, d(X), c(X), e(X), a \rangle$ | $25 + 2 \cdot 4(5 + 0.5(10 + 0.1 \cdot 10)) = 109$ |
| $\langle e(X), c(X), a, d(X), b \rangle$ | complete ordering | |

Table 3: A trace of a sample run of Algorithm 4 on the set of Figure 8. The left column shows the cheapest prefix extracted from the list on each step, the middle column – its extensions or completions that are added to the list, and the right column – their associated costs.

For comparison, we now show how the same task is performed by Algorithm 4. The algorithm maintains a list of prefixes, sorted by their cost values, and which initially contains an empty sequence. On each step the algorithm extracts from the list its cheapest element, and adds to the list the extensions or completions of this prefix. Extensions are created when the set of remaining subgoals is dependent, by appending each of the remaining subgoals to the end of the prefix. Completions are created when the set of remaining subgoals is independent, by sorting them and appending the entire resulting sequence to the prefix. An extension is added to the list only when the adjacency restriction test succeeds on its two last subgoals. To make the list operations faster, we can implement it as a heap structure (Cormen et al., 1991).

The trace of Algorithm 4 on the set $\mathcal{S}_0$ is shown in Table 3. The left column shows the cheapest prefix extracted from the list on each step, the middle column – its extensions or completions that are added to the list, and the right column – their associated costs.

It looks as if the DAC algorithm orders the given set $\mathcal{S}_0$ more efficiently than Algorithm 4. We can compare several discrete measurements to show this. For example, Algorithm 6 performs 4 sorting sessions, each one with 2 elements, while Algorithm 4 performs 5 sortings with 2 elements, and 3 sortings with 3 elements. The adjacency restriction is tested only 3 times by Algorithm 6, and 11 times by Algorithm 4. Algorithm 6 creates totally 8 different ordered sub-sequences, with total length 22, while Algorithm 4 creates 24 ordered prefixes, with total length 55.

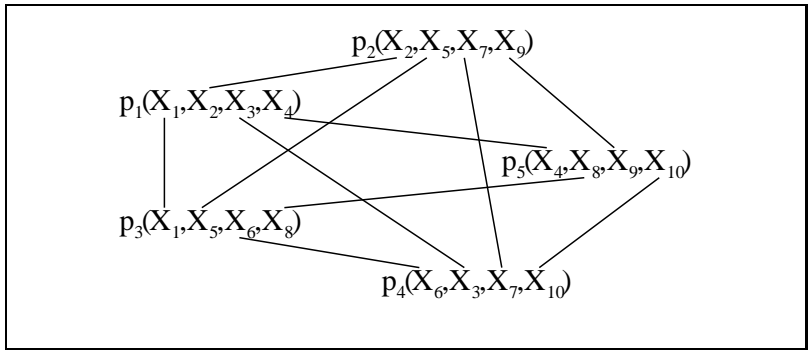

Figure 14: An example of the worst case for ordering. When all variables are initially free, every subset of subgoals is indivisible under the binding of the rest of subgoals, and the overall complexity of ordering by Algorithm 6 is $O(n!)$.

## 4.8 Complexity Analysis

Both Algorithm 4 and Algorithm 6 find a minimal ordering, and both sort independent subsets of subgoals whenever possible. Algorithm 6, however, offers several advantages due to its divide-and-conquer strategy.

Let $n$ be the number of subgoals in the initial set. For convenience, we assume that the time of computing the control values for one subgoal is $O(1)$; otherwise, if this time is $\tau$, all the complexities below must be multiplied by $\tau$. The worst case complexity of Algorithm 6 is $O(n!)$. Figure 14 shows an example of such a case for $n = 5$. In this set every two subgoals share a variable that does not appear in other subgoals. Thus, other subgoals cannot bind it. The set of the root is indivisible, and no matter which binder is chosen, the sets of the children are indivisible. So, in each child of the root, we must select every remaining subgoal as the binder, and so on. The overall complexity of this execution is $O(n!)$. This is indeed the worst-case complexity: presence of AND-nodes in the tree can only reduce it.

Note that even when $n$ is small, such a complex rule body with $\binom{n}{2}$ free variables is very improbable in practical programs. Also, the worst-case complexity can be reduced to $O(n^2 \times 2^n)$, if we move from divisibility trees to *divisibility graphs* (DAGs), where all identical nodes of a divisibility tree (same subgoal set, same binding set) are represented by a single vertex. The equivalence test of the tree nodes can be performed efficiently with the help of trie structures (Aho et al., 1987), where subgoals are sorted lexicographically.

Let there be $n$ subgoals, with $v$ shared variables appearing in $m$ subgoals. As was already noted in Section 4.3, the partition of subgoals into subsets can be performed in $O(n)$ average time, using a Union-Find data structure (Cormen et al., 1991, Chapter 22). In the worst possible case, there are no AND-nodes in the divisibility tree, apart from the root node (whose set is divisible into a dependent set of size $m$ and an independent set of size $n - m$). The overall complexity of the DAC algorithm in such a case is

$$
\begin{aligned}
T(n, m, v) \;=\; & O(n) & & \text{— divisibility partition} \\
+ \; & O((n - m)\log(n - m)) & & \text{— ordering of independent subgoals} \\
+ \; & O((\textstyle\prod_{i=0}^{k}(m - i)) \cdot \log(m - k)) & & \text{— ordering of dependent subgoals} \\
+ \; & O(m \cdot \textstyle\prod_{i=0}^{k-1}(m - i)) & & \text{— folding} \\
+ \; & O(n \cdot \textstyle\prod_{i=0}^{k-1}(m - i)) & & \text{— merging}
\end{aligned}
$$

where $k$ is the maximal possible number of bindings performed before the remaining subset is independent. If we assume that every subgoal binds all its free variables (which happens very frequently in practical logic programs), then $k = \min\{v, m - 1\}$; otherwise $k = m - 1$. $k$ is equal to the maximal number of OR-nodes on a path from the root to a leaf of the divisibility tree. Therefore, the height of the divisibility tree is limited by $k + 1$. Actually, the tree can be shallower, since some binders can bind more than one shared variable each. This means that the number of shared variables can decrease by more than 1 in each OR-node. Below we simplify the above formula for several common cases, when $k$ is small and when the abovementioned assumption holds (every subgoal binds all its free variables after its proof terminates).

- If $v < m \ll n$: $T(n, m, v) = O(n \cdot m^v + n \cdot \log n)$

- If $m \leq v \ll n$: $T(n, m, v) = O(n \cdot m^{m-1} + n \cdot \log n)$

- If $v \ll m \simeq n$: $T(n, m, v) = O(n^{v+1} \cdot \log n)$

- If $m \ll v \simeq n$: $T(n, m, v) = O(n \cdot m! + n \cdot \log n)$

Generally, for a small number $v$ of shared variables, the complexity of the algorithm is roughly bounded by $O(n^{v+1} \cdot \log n)$. In particular, if all subgoals are independent ($v = 0$), the complexity is $O(n \log n)$. In most practical cases, the number of shared free variables in a rule body is relatively small, and every subgoal binds all its free variables; therefore, the algorithm has polynomial complexity. Note that even if a rule body in the program text contains many free variables, most of them usually become bound after the rule head unification is performed (i.e., before we start the ordering of the instantiated body).

## 5. Learning Control Knowledge for Ordering

The ordering algorithms described in the previous sections assume the availability of correct values of average cost and number of solutions for various predicates under various argument bindings. In this section we discuss how this control knowledge can be obtained by learning.

Instead of static exploration of the program text (Debray & Lin, 1993; Etzioni, 1993), we adopt the approach of Markovitch and Scott (1989) and learn the control knowledge by collecting statistics on the literals that were proved in the past. This learning can be performed on-line or off-line. In the latter case, the ordering system first works with a *training set* of queries, while collecting statistics. This training set can be built on the distribution of user queries seen in the past. We assume that the distribution of queries received by the system does not change significantly with time; hence, the past distribution directs the system to learn relevant knowledge for the future queries.

While proving queries, the learning component accumulates information about the control values (average cost and number of solutions) of various literals. Storing a separate value for each literal is not practical, for two reasons. The first is the large space required by this approach. The second is the lack of generalization: the ordering algorithm is quite likely to encounter literals which have not been seen before, and whose control values are unknown. Recall that when we transformed Equation 2 into Equation 5, we moved from control values of single literals to average control values over *sets of literals*. To obtain the precise averages for these sets, we still needed the control values of individual literals. Here, we take a different approach, that of learning and using control values for *more general classes* of literals. The estimated $\overline{cost}$ ($\overline{nsols}$) value of a class can be defined as the average real cost (nsols) value of all examples of this class that were proven in the past.

The more refined the classes, the smaller the variance of real control values inside each class, the more precise the $\overline{cost}$ and $\overline{nsols}$ estimations that the classes assign to their members, and the better orderings we obtain. One easy way to define classes is by *modes* or *binding patterns* (Debray & Warren, 1988; Ullman & Vardi, 1988): for each argument we denote whether it is free or bound. For example, for the predicate `father` the possible classes are `father(free,free)`, `father(bound,free)`, `father(free,bound)` and `father(bound,bound)`. Now, if we receive a literal (for example, `father(abraham,X)`), we can easily determine its binding pattern (in this case, `father(bound,free)`) and retrieve the control information stored for this class. Of course, to find the binding pattern of a subgoal with a given binding set, we need a method to determine which variables are bound by the subgoals of the binding set. The same problem arose in *DPart* computation (Section 4.3). We shall discuss some practical ways to solve this problem in Section 7.1.

For the purpose of class definition we can also use *regression trees* – a type of decision tree that classifies to continuous numeric values and not to discrete classes (Breiman et al., 1984; Quinlan, 1986). Two separate regression trees can be stored for every program predicate, one for its $\overline{cost}$ values, and one for the $\overline{nsols}$. The tests in the tree nodes can be defined in various ways. If we only use the test *"is argument i bound?"*, then the classes of literals defined by regression trees coincide with the classes defined by binding patterns. But we can also apply more sophisticated tests, both syntactic (e.g., *"is the third argument a term with functor* `f`*?"*) and semantic (e.g., *"is the third argument female?"*), which leads to more refined classes and better estimations. A possible regression tree for estimating the of number of solutions for predicate `father` is shown in Figure 15.

Semantic tests about the arguments require logic inference (in the example of Figure 15 – invoking the predicate `female` on the first argument of the literal). Therefore, they must be as efficient as possible. Otherwise the retrieval of control values will take too much time. The problem of efficient learning of control values is further considered elsewhere (Ledeniov & Markovitch, 1998a).

Several researchers applied machine learning techniques for accelerating logic inference (Cohen, 1990; Dejong & Mooney, 1986; Langley, 1985; Markovitch & Scott, 1993; Minton, 1988; Mitchell, Keller, & Kedar-Cabelli, 1986; Mooney & Zelle, 1993; Prieditis & Mostow, 1987). Some of these works used explanation-based learning or generalized caching tech-niques to avoid repeated computation. Others utilized the acquired knowledge for the problem of *clause selection*. None of these works, however, dealt with the problem of subgoal reordering.

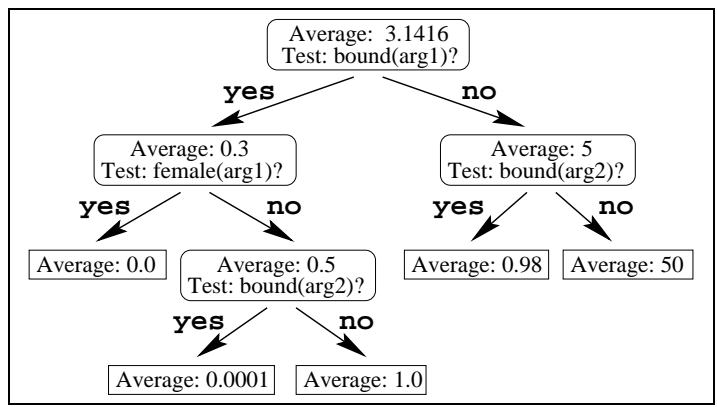

Figure 15: A regression tree that estimates the number of solutions for `father(arg1,arg2)`.

## 6. Experimentation

To test the effectiveness of our ordering algorithm, we experimented with it on various domains, and compared its performance to other ordering algorithms. Most experiments were performed on randomly created artificial domains. We also tested the performance of the system on several real domains.

### 6.1 Experimental Methodology

All experiments described below consist of a training session, followed by a testing session. Training and testing sets of queries are randomly drawn from a fixed distribution. In the training session we collect the control knowledge for literal classes. In the testing session we prove the queries of the testing set using different ordering algorithms, and compare their performance using various measurements.

The goal of ordering is to reduce the time spent by the Prolog interpreter when it proves queries of the testing set. This time is the sum of the time spent by the ordering procedure (*ordering time*) and the time spent by the interpreter (*inference time*). Since the CPU time is known to be very sensitive to irrelevant factors such as hardware, software and programming quality, we also show two alternative discrete measurements: the total number of clause *unifications*, and the total number of clause *reductions* performed. The number of reductions reflects the size of the proof tree.

For experimentation we used a new version of the LASSY system (Markovitch & Scott, 1989), using regression trees for learning, and the ordering algorithms discussed in this paper.

## 6.2 Experiments with Artificial Domains

In order to ensure the statistical significance of the results of comparing different ordering algorithms, we experimented with many different domains. For this purpose, we created a set of 100 artificial domains, each with a small fixed set of predicates, but with a random number of clauses in each predicate, and with random rule lengths. Predicates in the rule bodies, and arguments in both rule heads and bodies are randomly drawn from fixed distributions. Each domain has its own training and testing sets (these two sets do not intersect).

The more training examples are fed into the system on the learning phase, the better estimations of control values it produces. On the other hand, the learning time must be limited, because after seeing a certain number of training examples, new examples do not bring much new information, and additional learning becomes wasteful. We have experimentally built a learning curve which shows the dependence of the quality of the control knowledge on the amount of training. The curve suggests that after control values were learned for approximately 400 literals, there is no significant improvement in the quality of ordering with new training examples. Therefore, in the subsequent experiments we stopped training after 600 cost values were learned. The training time was always small: one learned cost value corresponds to a complete proof of a literal. Thus, if every predicate in a program has four clauses that define it, then 600 cost values are learned after 2400 unifications, which is a very small time.

The control values were learned by means of regression trees (Section 5), with simple syntactic tests that only checked whether some argument is bound or whether some argument is a term with a certain functor (the list of functors was created automatically when the domain was loaded). However, as we shall see, even these simple tests succeeded in making good estimations of control values.

We tested the following ordering methods:

- **Random:** The subgoals are permuted randomly and the control knowledge is not used.

- **Algorithm 3:** Building ordered prefixes. Out of all prefixes that are permutation of one another, only the cheapest one is retained.

- **Algorithm 3a:** As Algorithm 3, but with best-first search method used to define the next processed prefix. A similar algorithm was used in the LASSY system of Markovitch and Scott (1989).

- **Algorithm 3b:** As Algorithm 3a, but with adjacency restriction test added. A similar algorithm was described by Smith and Genesereth (1985).

- **Algorithm 4:** As Algorithm 3b, but whenever all the subgoals that are not in the prefix are independent (under the binding of the prefix), they are sorted and the result is appended to the prefix as one unit.

- **Algorithm 6:** The DAC algorithm.

In our experiments we always used the Bubble-Sort algorithm to sort literals in independent sets. This algorithm is easy to implement, and it is known to be efficient for small sets, when the elements are already ordered, or nearly ordered. In practice, programmers order most program rules optimally, and the sorting stops early.

| Ordering Method | Unifications | Reductions | Ordering Time | Inference Time | Total Time | Ord.Time Reductions |
|---|---|---|---|---|---|---|
| Random | 86052.06 | 27741.52 | 8.1910 | 27.385 | 35.576 | 0.00029 |
| Algorithm 3 | 2600.32 | 911.04 | 504.648 | 1.208 | 505.856 | 0.55 |
| Algorithm 3.a | 2829.00 | 978.59 | 347.313 | 1.178 | 348.491 | 0.35 |
| Algorithm 3.b | 2525.34 | 881.12 | 203.497 | 1.137 | 204.634 | 0.23 |
| Algorithm 4 | 2822.27 | 976.02 | 40.284 | 1.191 | 41.475 | 0.04 |
| Algorithm 6 | 2623.82 | 914.67 | 2.3620 | 1.102 | 3.464 | 0.0025 |

Table 4: The effect of ordering on the tree sizes and the CPU time (mean results over 100 artificial domains).

Since the non-deterministic nature of the random method introduces additional noise, we performed on each artificial domain 20 experiments with this method, and the table presents the average values of these measurements.

Table 4 shows the obtained results over 100 domains: the rows correspond to the ordering methods used, and the columns to the measurements taken. The rightmost column shows the ratio of the ordering time and the number of reductions performed, which reflects the average ordering time of one rule body. The inference time was not measured separately, but was set as the difference of the total time and the ordering time.

Several observations can be made:

1. Using the DAC ordering algorithm helps to reduce the total time of proving the testing set of queries by a factor of 10, compared to the random ordering. The inference time is reduced by a factor of 25.

2. All deterministic ordering methods have similar number of unifications and reductions, and similar inference time, which is predictable, since they all find minimal orderings. Small fluctuations of these values can be explained by the fact that some rules have several minimal orderings under the existing control knowledge, and different ordering algorithms select different minimal orderings. Since the control knowledge is not absolutely precise, the real execution costs of these orderings may be different, which leads to the differences. The random ordering method builds much larger trees, with larger inference time.

3. When we compare the performance of the deterministic algorithms $(3 - 6)$, we see that the DAC algorithm performs much better than the algorithms that build ordered prefixes. In the latter ones, the ordering is expensive, and smaller inference time cannot compensate for the increase in ordering time. Only Algorithm 4, a combination of several ideas of previous researchers, has total time comparable with the time of the random method (though still greater).

4. It may seem strange that the simple random ordering method has larger ordering time than the sophisticated Algorithm 6. To explain this, note that the random method creates much larger proof trees (on average), therefore the number of ordered rules increases, and even the cheap operations, like random ordering of a rule, sum up to a considerable time. The average time spent on ordering of one rule is shown in the last column of Table 4; this value is very small in the random method.

## 6.3 Experiments with Real Domains

We tested our ordering algorithm also on real domains obtained from various sources. These domains allow us to compare orderings performed by our algorithm with orderings performed by human programmers.

The following domains were used:

- **Moral-reasoner:** Taken from the Machine Learning Repository at the University of California, Irvine[1]. The domain qualitatively simulates moral reasoning: whether a person can be considered guilty, given various aspects of his character and of the crime performed.

- **Depth-first planner:** Program 14.11 from the book "The Art of Prolog" (Sterling & Shapiro, 1994). The program implements a simple planner for the blocks world.

- **Biblical Family Database:** A database similar to that described in Example 1.

- **Appletalk:** A domain describing the physical layout of a local computer network (Markovitch, 1989).

- **Benchmark:** A Prolog benchmark taken from the CMU Artificial Intelligence Repository[2]. The predicate names are not informative: it is an example of a program where manual ordering is difficult.

- **Slow reverse:** Another benchmark program from the same source.

- **Geography:** Also a benchmark program from the CMU Repository. The domain contains many geographical facts about countries.

Table 5 shows the results obtained. For ordering we used the DAC algorithm, with literal classes defined by binding patterns. It can be seen that the DAC algorithm was able to speed up the logic inference in real domains as well. Note that in the Slow Reverse domain the programmer's ordering was already optimal; thus, applying the ordering algorithm did not reduce the tree sizes. Still, the overhead of the ordering is not significant.

## 7. Discussion

In this concluding section we discuss several issues concerning the practical implementation of the DAC algorithm and several ways to increase its efficiency. Then we survey some related areas of logic programming and propose the use of the DAC algorithm there.

---

1. URL: http://www.ics.uci.edu/~mlearn/MLRepository.html
2. URL: http://www.cs.cmu.edu/afs/cs.cmu.edu/project/ai-repository/ai/html/air.html

| Domain | Without ordering | | With ordering | | Gain ratio |
|---|---|---|---|---|---|
| | unifications | seconds | unifications | seconds | (time/time) |
| Moral-reasoner | 352180 | 98.39 | 87020 | 23.53 | 4.2 |
| Depth-first planner | 10225 | 19.01 | 9927 | 18.16 | 1.05 |
| Biblical Family | 347827 | 112.68 | 120701 | 46.08 | 2.5 |
| Appletalk | 5036167 | 1246.30 | 640092 | 221.73 | 5.6 |
| Benchmark | 62012 | 554.31 | 46012 | 395.04 | 1.4 |
| Slow reverse | 6291 | 10.33 | 6291 | 11.92 | 0.9 |
| Geography | 428480 | 141.47 | 226628 | 82.76 | 1.7 |

Table 5: Experiments on real domains.

## 7.1 Practical Issues

In this subsection we would like to address several issues related to implementation and applications of the DAC algorithm.

The computation of the *DPart* function (Section 3.2.1) requires a procedure for computing the set of variables bound by a given binding set of subgoals. The same procedure is needed for computing control values (Section 5). There are several possible ways to implement such a procedure. For example:

1. The easiest way is to *assume* that every subgoal binds all the variables appearing in its arguments. This simplistic assumption is sufficient for many domains, especially the database-oriented ones. However, it is not appropriate when logic programs are used to manipulate complex data structures containing free variables (such as difference lists). This assumption was used for the experiments described in Section 6.

2. Some dialects of Prolog and other logic languages support *mode declarations* provided by the user (Somogyi et al., 1996b). When such declarations are available, it is easy to infer the binding status of each variable upon exiting a subgoal.

3. Even when the user did not supply enough mode declarations, they can often be inferred from the structure of the program by means of *static analysis* (Debray & Warren, 1988). Note, however, that as was pointed out by Somogyi et al. (1996b), no-one has yet demonstrated a mode inference algorithm that is guaranteed to find accurate mode information for every predicate in the program.

4. We can *learn* the sets of variables bound by classes of subgoals using methods similar to those described in Section 5 for learning control values.

Several researchers advocate user declarations of available (permitted) modes. Such declarations can be elegantly incorporated into our algorithm to prune branches that violate available modes. When we fix a binder in an OR-node, we compute the set of variables that become bound by it. If this results in a violation of an available mode for one of the subgoals of the corresponding child, then the whole subtree of this child is pruned. Note that we can detect violations even when the mode of the subgoal is partially unknown

---

**CandidateSet**$(\mathcal{S}, \mathcal{B})$

    let $\{\mathcal{S}_1, \mathcal{S}_2, \ldots \mathcal{S}_k\} \leftarrow DPart(\mathcal{S}, \mathcal{B})$

    case

       $\ldots$

       • $k = 1$, *shared-vars*$(\mathcal{S}_1) \neq \emptyset$ ($\mathcal{S}$ is <u>indivisible</u> under $\mathcal{B}$):

          loop for $A \in \mathcal{S}$

             if $\mathcal{B} \cup \{A\}$ does not violate available modes

                in any subgoal of $\mathcal{S} \setminus \{A\}$

             then

                let $\mathcal{C}(A) \leftarrow$ **CandidateSet**$(\mathcal{S} \setminus \{A\}, \mathcal{B} \cup \{A\})$

                let $\mathcal{C}'(A) \leftarrow \left\{ \textbf{Fold}(A \| \vec{O}_A, \mathcal{B}) \mid \vec{O}_A \in \mathcal{C}(A) \right\}$

             else let $\mathcal{C}'(A) \leftarrow \emptyset$   (don't enter the branch)

          Return $\bigcup_{A \in \mathcal{S}} \mathcal{C}'(A)$

---

Figure 16: Changes to Algorithm 6 that make use of available mode declarations. The rest of the algorithm remains unchanged.

at the moment. For example, if all the available modes require that the first argument be unbound, then binding of the argument by the OR-node binder will trigger pruning, even if the binding status of the other arguments is not yet known. Figure 16 shows how Algorithm 6 can be changed in order to incorporate declarations of available modes. Any other correctness requirement can be treated in a similar manner: a candidate ordering will be rejected whenever we see that it violates the requirement.

The experiments described in Section 6 were performed with a Prolog interpreter. Is it possible to combine the DAC algorithm with a Prolog compiler? There are several ways to achieve this goal. One way is to allow the compiler to insert code for on-line learning. The compiled code will contain procedures for accumulating control values and for the DAC algorithm. Alternatively, off-line learning can be implemented, with training as a part of the compilation process.

Another method for combining our algorithm with existing Prolog compilers is to use it for program transformation, and to process the transformed program by a standard compiler. Elsewhere (Ledeniov & Markovitch, 1998a) we describe the method for classifying the orderings produced by the DAC algorithm. For each rule we build a classification tree, where classes are the different orderings of the rule body, and the tests are applied to the rule head arguments. These are the same type of tests described in Section 5 for learning control values. Figure 17 shows two examples of such trees.

Given such a classification tree, we can write a set of Prolog rules, where each rule has the same head as the original rule, and has a body built of all the tests on the path from the tree root to a leaf node followed by the ordering at the leaf. For example, the second tree in Figure 17 yields the following set of rules:

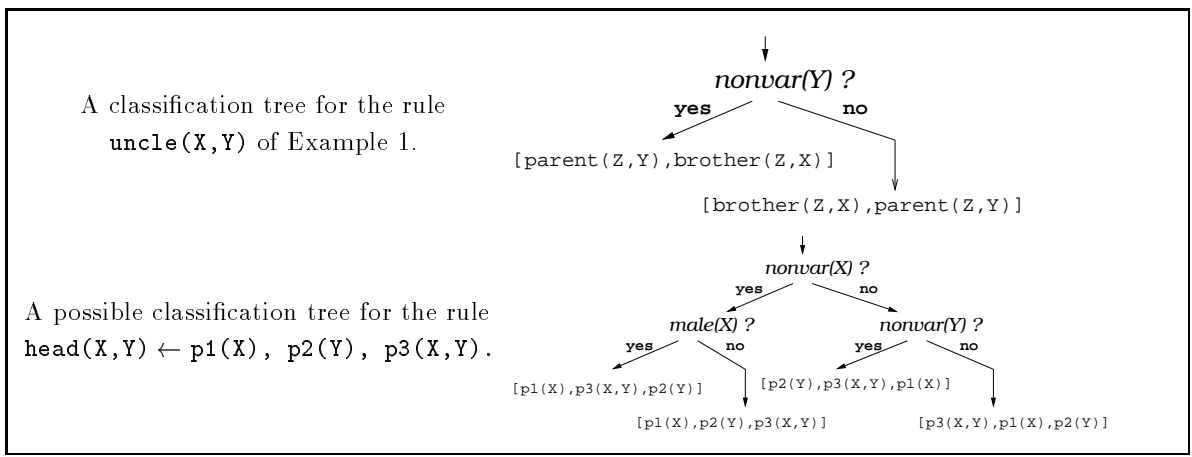

Figure 17: Examples of classification trees that learn rule body orderings.

```
head(X,Y) ← nonvar(X), male(X), p1(X), p3(X,Y), p2(Y).
head(X,Y) ← nonvar(X), not(male(X)), p1(X), p2(Y), p3(X,Y).
head(X,Y) ← var(X), nonvar(Y), p2(Y), p3(X,Y), p1(X).
head(X,Y) ← var(X), var(Y), p3(X,Y), p1(X), p2(Y).
```

From Table 4 we can see that while the DAC algorithm helped to reduce the inference time by a factor of 25, the total time was reduced only by a factor of 10. This difference is caused by the additional computation of the ordering procedure. There is a danger that the benefit obtained by ordering will be outweighed by the cost of the ordering process. This is a manifestation of the so-called *utility problem* (Minton, 1988; Markovitch & Scott, 1993). In systems that are strongly-moded (such as Mercury – Somogyi et al., 1996b) we can employ the DAC algorithm statically at compilation time for each one of the available modes, thus reducing the run-time ordering time to zero. The mode-based approach performs only syntactic tests of the subgoal arguments. The classification tree method, described above, is a generalization of the mode-based approach, allowing semantic tests as well.

Due to insufficient learning experience or lack of meaningful semantic tests, it is quite possible that the classification trees contain leaves with large degrees of error. In such cases we still need to perform the ordering dynamically. To reduce the harmfulness of the utility problem in the case of dynamic ordering, we can use a cost-sensitive variation of the DAC algorithm (Ledeniov & Markovitch, 1998a, 1998b). This modified algorithm deals with the problem by explicit reasoning about the economy of the control process. The algorithm is *anytime*, that is, it can be stopped at any moment and return its currently best ordering (Boddy & Dean, 1989). We learn a *resource-investment function* to compute the expected return in speedup time for additional control time. This function is used to determine a stopping condition for the anytime procedure. We have implemented this framework and found that indeed we have succeeded in reducing ordering time, without significant increase of inference time.

## 7.2 Relationship to Other Works

The work described in this paper is a continuation of the line of research initiated by Smith and Genesereth (1985) and continued by Natarajan (1987) and Markovitch and Scott (1989). This line of research aims at finding the most efficient ordering of a set of subgoals. The search for minimal-cost ordering is based on cost analysis that utilizes available information about the cost and number-of-solutions of individual subgoals.

Smith and Genesereth (1985) performed an exhaustive search over the space of all permutations of the given set of subgoals, using the *adjacency restriction* to reduce the size of the search space (Equation 8). This restriction was applied on pairs of adjacent subgoals in the global ordering of the entire set. When applied to an *independent set of subgoals*, the adjacency restriction is easily transformed into the sorting restriction: the subgoals in a minimal ordering must be sorted by their *cn* values. Natarajan (1987) arrived at this conclusion and presented an efficient ordering algorithm for independent sets.

The DAC algorithm uses subgoal dependence to break the set into smaller subsets. Independent subsets are sorted. Dependent subsets are recursively ordered, and the resulting orderings are merged using a generalization of the adjacency restriction that manipulates blocks of subgoals. Therefore the DAC algorithm is a generalization of both algorithms.

During the last decade, a significant research effort went into static analysis (SA) of logic programs. There are three types of SA that can be exploited by the DAC algorithm to reduce the ordering time.

A major part of the SA research deals with *program termination* (De Schreye & Decorte, 1994). The DAC algorithm solves the termination problem, as a special case of the efficiency problem (it always finds a terminating ordering, if such orderings exist). During learning, we set limits on the computation resources available for subgoal execution. If a subgoal is non-terminating (in a certain mode), the learning module will associate a very high cost with this particular mode. Consequently, the DAC algorithm will not allow orderings with this mode of the subgoal. Nevertheless, while the use of static termination analysis is not mandatory for a proper operation of the DAC algorithm, we can exploit such analysis to increase the efficiency of both the learning process and the ordering process. During learning, the limit that we set on the computation resources devoted to the execution of a subgoal must be high, to increase the reliability of the cost estimation. However, such a high limit can lead to a significant increase in learning time when many subgoals are non-terminating. If termination information obtained by SA is available, we can use it to avoid entering infinite branches of proof trees. During ordering, termination information can serve to reduce the size of space of orderings searched by the algorithm. If the termination information comes in the form of allowed modes (Somogyi et al., 1996b), orderings that violate these modes are filtered out, as in the modified algorithm shown in Figure 16. If the termination information comes in the form of a partial order between subgoals, orderings that violate this partial order can be filtered out in a similar manner.

The second type of SA research that can be combined with the DAC algorithm is *correctness analysis*, where the program is tested against specifications given by the user.

The FOLON environment (Henrard & Le Charlier, 1992) was designed to support the methodology for logic program construction that aims at reconciling the declarative semantics with an efficient implementation (Deville, 1990). The construction process starts with a specification, converts it into a logic description and finally, into a Prolog program. If the rules of the program are not correct with respect to the initial specification, the system performs transformations such as reordering literals in a clause, adding type checking literals and so on. De Boeck and Le Charlier (1990) mention this reordering, but do not specify an ordering algorithm different from the simple generate-and-test method. Cortesi, Le Charlier, and Rossi (1997) present an analyzer for verifying the correctness of a Prolog program relative to a specification which provides a list of input/output annotations for the arguments and parameters that can be used to establish program termination. Again, no ordering algorithm is given explicitly. The purpose of the DAC algorithm is complementary to the purpose of FOLON, and it could serve as an auxiliary aid to make the resulting Prolog program more efficient.

Recently, the Mercury language was developed at the University of Melbourne (Somogyi et al., 1996a, 1996b). Mercury is a strongly typed and strongly moded language. Type and mode declarations should be supplied by the programmer (though recent releases of the Mercury system already support partial inference of types and modes – Somogyi et al., 1996a). The compiler checks that mode declarations for all predicates are satisfied; if necessary, it *reorders* subgoals in the rule body to ensure mode correctness (and rejects the program if neither ordering satisfies the mode declaration constraints). When the compiler performs this reordering, it does not consider the efficiency issue. It often happens that several orderings of a rule body satisfy the mode declaration constraints: in such cases the Mercury compiler could call the static version of the DAC algorithm to select the most efficient ordering. Another alternative is to augment the DAC algorithm by mode declaration checks, as was shown in Figure 16.

Note that Mercury is a purely declarative logic programming language, and is therefore more suitable for subgoal reordering than Prolog. It has no non-logical constructs that could destroy the declarative semantics which give logic programs their power; in Mercury even I/O is declarative.

The third type of relevant SA is the *cost analysis* of logic programs (Debray & Lin, 1993; Braem et al., 1994; Debray et al., 1997). Cortesi et al. (1997) describe a cost formula similar to Equation 5 to select a lowest-cost ordering. However, they used a generate-and-test approach which can sometimes be prohibitively expensive. Static analysis of cost and number of solutions can be used to obtain the control values, instead of learning them.

The efficiency of logic programs can also be increased by methods of *program transformation* (Pettorossi & Proietti, 1994, 1996). One of the most popular approaches is the "rules+strategies" approach, which consists in starting from an initial program and then applying one or more elementary *transformation rules*. Transformation strategies are meta-rules which prescribe suitable sequences of applications of transformation rules.

One of the possible transformation rules is the *goal rearrangement rule* which transforms a program by transposing two adjacent subgoals in a rule body. Obviously, any ordering of a rule body can be transformed into any other ordering by a finite number of such transpositions. Thus, static subgoal ordering can be considered a special case of program transformation where only the goal rearrangement rule is used. On the other hand, dynamic and semi-dynamic ordering methods cannot be represented by simple transformation rules, since they make use of run-time information (expressed in bindings that rule body subgoals obtain through unifications of rule heads), and may order the same rule body differently under different circumstances.

A program transformation technique called *compiling control* (Bruynooghe, De Schreye, & Krekels, 1989; Pettorossi & Proietti, 1994) follows an approach different from that of trying to improve the control strategy of logic programs. Instead of enhancing the naive Prolog evaluator using a better (and often more complex) computation rule, the program is transformed so that the derived program behaves under the naive evaluator exactly as the initial program would behave under an enhanced evaluator. Most forms of compiling control first translate the initial program into some standard representation (for example, into an *unfolding tree*), while the complex computation rule is used, and then the new program is constructed from this representation, with the naive computation rule in mind.

Reordering of rule body subgoals can be regarded as moving to a complex computation rule which selects subgoals in the order dictated by the ordering algorithm. In the case of the DAC algorithm, this computation rule may be too complex for simple use of compiling control methods. Nevertheless, it can be easily incorporated into a special compiling control method. In Section 7.1 we described a method of program rewriting which first builds classification trees based on the orderings that were performed in the past, and then uses these classification trees for constructing clauses of a derived program. The derived program can be efficiently executed under the naive computation rule of Prolog. This technique is in fact a kind of compiling control. Its important property is the use of knowledge collected from experience (the orderings that were made in the past).

One transformation method that can significantly benefit from the DAC algorithm is *unfolding* (Tamaki & Sato, 1984). During the unfolding process subgoals are replaced by their associated rule bodies. Even if the initial rules were ordered optimally by a human programmer or a static ordering procedure, the resulting combined sequence may be far from optimal. Therefore it could be very advantageous to use the DAC algorithm for reordering of the unfolded rule. As the rules become longer, the potential benefit of ordering grows. The danger of high complexity of the ordering procedure can be overcome by using the cost-sensitive version of the DAC algorithm (Section 7.1).

## 7.3 Conclusions

In this work we study the problem of subgoal ordering in logic programs. We present both a theoretical base and a practical implementation of the ideas, and show empirical results that confirm our theoretical predictions. We combine the ideas of Smith and Genesereth (1985), Simon and Kadane (1975) and Natarajan (1987) into a novel algorithm for ordering of conjunctive goals. The algorithm is aimed at minimizing the time which the logic interpreter spends on the proof of the given conjunctive goal.

The main algorithm described in this paper is the DAC algorithm (Algorithm 6, Section 4.6). It works by dividing the sets of subgoals into smaller sets, producing candidate sets of orderings for the smaller sets, and combining these candidate sets to obtain orderings of the larger sets. We prove that the algorithm finds a minimal ordering of the given set of subgoals, and we show its efficiency under practical assumptions. The algorithm can be employed statically (to reorder rule bodies in the program text before the execution starts), semi-dynamically (to reorder the rule body before the reduction is performed) or dynamically (to reorder the resolvent after every reduction of a subgoal by a rule body).

Several researchers (Minker, 1978; Warren, 1981; Naish, 1985a, 1985b; Nie & Plaisted, 1990) proposed various heuristics for subgoal ordering. Though fast, these methods do not guarantee finding minimal-cost orderings. Our algorithm provably finds a minimal-cost ordering, though the ordering itself may take more time than with the heuristic methods. In the future it seems promising to incorporate heuristics into the DAC algorithm. For example, heuristics can be used to grade binders in OR-nodes: rather than exhaustively trying all subgoals as binders, we could try just one, or several binders, thus reducing the ordering time. Also, the current version of our ordering algorithm is suitable only for finding *all* solutions to a conjunctive goal. We would like to extend it to the problem of finding one solution, or a fixed number of solutions.

Another interesting issue for further research is the adaptation of the DAC algorithm to interleaving ordering methods (Section 2.3). There, if subgoals of a rule body are added to an ordered resolvent, it seems wasteful to start a complete ordering process; we should use the information stored in the existing ordering of the resolvent. Perhaps the whole divisibility tree of the resolvent should be stored, and its nodes updated when subgoals of a rule body are added to the resolvent.

The ordering algorithm needs control knowledge for its work. This control knowledge is the average cost and number of solutions of literals, and it can be learned by training and collecting statistics. We make an assumption that the distribution of queries received by the system does not change with time; thus, if the training set is based on the distribution seen in the past, the system learns relevant knowledge for future queries. We consider the issue of learning control values more thoroughly in another paper (Ledeniov & Markovitch, 1998a), together with other issues concerning the DAC algorithm (such as minimizing the total time, instead of minimizing the inference time only).

Ullman and Vardi (1988) showed that the problem of ordering subgoals to obtain termination is inherently exponential in time. The problem we work with is substantially harder: we must not only find an order whose execution terminates in finite time, but one that terminates in minimal finite time. It is impossible to find an efficient algorithm for all cases. The DAC algorithm, however, is efficient in most practical cases, when the graph representing the subgoal dependence (Figure 3) is sparsely connected.

We have implemented the DAC algorithm and tested it on artificial and real domains. The experiments show a speedup factor of up to 10 compared with random ordering, and up to 13 compared with some alternative ordering algorithms.

The DAC algorithm can be useful for many practical applications. Formal hardware verification has become extremely important in the semiconductor industry. While model checking is currently the most widely used technique, it is generally agreed that coping with the increasing complexity of VLSI design requires methods based on theorem proving. The main obstacle preventing the use of automatic theorem proving is its high computational demands. The DAC algorithm may be used for speeding up logic inference, making the use of automatic theorem provers more practical.

Logic has gained increasing popularity for representation of common-sense knowledge. It has several advantages, including flexibility and well-understood semantics. Indeed, the CYC project (Lenat, 1995) has recently moved from frame-based representation to logic-based representation. However, the large scale of such knowledge bases is likely to present significant efficiency problems to the inference engines. Using automatic subgoal ordering techniques, such as those described here, may help to solve these problems.

The issue of subgoal ordering obtains a new significance with the development of Inductive Logic Programming (Lavrac & Dzeroski, 1994; Muggleton & De Raedt, 1994). Systems using this approach, such as FOIL (Quinlan & Cameron-Jones, 1995), try to build correct programs as fast as possible, without considering the efficiency of the produced programs. Combining the DAC algorithm with Inductive Logic Programming and other techniques for the synthesis of logic programs (such as the deductive and the constructive approaches) looks like a promising direction.

## Appendix A. Proof of Lemma 8

In this appendix we present the proof of a lemma which was omitted from the main text of the paper for reasons of compactness. Before we prove it we show two auxiliary lemmas.

### Lemma 9
*Let $\vec{A}_1$ and $\vec{A}_2$ be two ordered sequences of subgoals, and $\mathcal{B}$ a set of subgoals. The value of $cn(\vec{A}_1\|\vec{A}_2)|_{\mathcal{B}}$ lies between the values $cn(\vec{A}_1)|_{\mathcal{B}}$ and $cn(\vec{A}_2)|_{\mathcal{B}\cup\vec{A}_1}$.*

### Proof:
Denote
$$
\begin{aligned}
c_1 &= c\overline{os}t(\vec{A}_1)|_{\mathcal{B}} & n_1 &= n\overline{so}ls(\vec{A}_1)|_{\mathcal{B}} & cn_1 &= cn(\vec{A}_1)|_{\mathcal{B}} \\
c_2 &= c\overline{os}t(\vec{A}_2)|_{\mathcal{B}\cup\vec{A}_1} & n_2 &= n\overline{so}ls(\vec{A}_2)|_{\mathcal{B}\cup\vec{A}_1} & cn_2 &= cn(\vec{A}_2)|_{\mathcal{B}\cup\vec{A}_1} \\
c_{1,2} &= c\overline{os}t(\vec{A}_1\|\vec{A}_2)|_{\mathcal{B}} & n_{1,2} &= n\overline{so}ls(\vec{A}_1\|\vec{A}_2)|_{\mathcal{B}} & cn_{1,2} &= cn(\vec{A}_1\|\vec{A}_2)|_{\mathcal{B}}
\end{aligned}
$$

$$
\begin{aligned}
cn_{1,2} &= \frac{n_{1,2}-1}{c_{1,2}} = \frac{n_1 n_2 - 1}{c_{1,2}} = \frac{(n_1-1)+n_1(n_2-1)}{c_{1,2}} = \\
&= \frac{c_1\frac{n_1-1}{c_1}+n_1 c_2\frac{n_2-1}{c_2}}{c_{1,2}} = \frac{c_1 cn_1 + n_1 c_2 cn_2}{c_{1,2}} = \frac{c_1}{c_{1,2}}\cdot cn_1 + \frac{n_1 c_2}{c_{1,2}}\cdot cn_2
\end{aligned}
$$

So, $cn_{1,2}$ always lies between $cn_1$ and $cn_2$ (because $\frac{c_1}{c_{1,2}}$ and $\frac{n_1 c_2}{c_{1,2}}$ are positive and sum to 1). More exactly, the point $cn_{1,2}$ divides the segment $[cn_1, cn_2]$ with ratio

$$(cn_{1,2}-cn_1) : (cn_2 - cn_{1,2}) = n_1 c_2 : c_1.$$

In other words, $cn_{1,2}$ is a *weighted average* of $cn_1$ and $cn_2$. Note that $c_1$ is the amount of resources spent in the proof-tree of $\vec{B}_1$, $n_1 c_2$ – the resources spent in the tree of $\vec{B}_2$, and $c_{1,2}$ is their sum. So, the more time (relatively) we dedicate to the proof of $\vec{B}_1$, the closer $cn_{1,2}$ is to $cn_1$. This conclusion can be generalized to a larger number of components in a concatenation (the proof is by induction):

$$
\begin{aligned}
cn(\vec{A}_1\|\vec{A}_2\|\ldots\vec{A}_k)|_{\mathcal{B}} &= \frac{c\overline{os}t(\vec{A}_1)|_{\mathcal{B}}}{c\overline{os}t(\vec{A}_1\|\vec{A}_2\|\ldots\vec{A}_k)|_{\mathcal{B}}}\cdot cn(\vec{A}_1)|_{\mathcal{B}} + \\
&+ \frac{n\overline{so}ls(\vec{A}_1)|_{\mathcal{B}}\cdot c\overline{os}t(\vec{A}_2)|_{\mathcal{B}\cup\vec{A}_1}}{c\overline{os}t(\vec{A}_1\|\vec{A}_2\|\ldots\vec{A}_k)|_{\mathcal{B}}}\cdot cn(\vec{A}_2)|_{\mathcal{B}\cup\vec{A}_1} + \ldots + \\
&+ \frac{n\overline{so}ls(\vec{A}_1\|\vec{A}_2\|\ldots\vec{A}_{k-1})|_{\mathcal{B}}\cdot c\overline{os}t(\vec{A}_k)|_{\mathcal{B}\cup\vec{A}_1\cup\ldots\vec{A}_{k-1}}}{c\overline{os}t(\vec{A}_1\|\vec{A}_2\|\ldots\vec{A}_k)|_{\mathcal{B}}}\cdot cn(\vec{A}_k)|_{\mathcal{B}\cup\vec{A}_1\cup\ldots\vec{A}_{k-1}}
\end{aligned}
$$

□

**Lemma 10**
Let $\mathcal{S}_0$ be a set of subgoals and $N$ be a node in the divisibility tree of $\mathcal{S}_0$. Let $\vec{O}_N = \vec{Q}\|\vec{A}_1\|\vec{A}_2\|\vec{R}$ be an ordering of $\mathcal{S}(N)$, where $\vec{A}_1$ and $\vec{A}_2$ are cn-equal max-blocks: $cn(\vec{A}_1)|_{\mathcal{B}(N)\cup\vec{Q}} = cn(\vec{A}_2)|_{\mathcal{B}(N)\cup\vec{Q}\cup\vec{A}_1}$.

Let $M$ be an ancestor of $N$ and $\vec{O}_M$ be an ordering of $\mathcal{S}(M)$ consistent with $\vec{O}_N$, where $\vec{A}_1$ and $\vec{A}_2$ are not violated. Then either $\vec{A}_1$ and $\vec{A}_2$ are both max-blocks in $\vec{O}_M$ and all max-blocks that stand between them are cn-equal to them, or $\vec{A}_1$ and $\vec{A}_2$ belong to the same max-block in $\vec{O}_M$, or $\vec{O}_M$ is MC-contradicting.

**Proof:** By induction on the distance between $N$ and $M$. If $M = N$, then $\vec{A}_1$ and $\vec{A}_2$ are max-blocks, and the lemma holds. Let $M \neq N$, and let $M'$ be the child of $M$ whose descendant is $N$. By inductive hypothesis, the lemma holds for $N$ and $M'$. Let $\vec{O}'_M$ be the projection of $\vec{O}_M$ on $M'$. $\vec{A}_1$ and $\vec{A}_2$ are not violated in $\vec{O}'_M$, since they are not violated in $\vec{O}_M$.

- If $\vec{A}_1$ and $\vec{A}_2$ are both max-blocks in $\vec{O}'_M$, then by the inductive hypothesis all max-blocks that stand between them are cn-equal to them. If $M$ is an OR-node, no new subgoals can enter between $\vec{A}_1$ and $\vec{A}_2$. If $M$ is an AND-node, the insertion of new subgoals is possible, but if it violates blocks, or places max-blocks not ordered by $cn$, then $\vec{O}_M$ is MC-contradicting, by Corollary 3 or Lemma 6. So, if $\vec{O}_M$ is not MC-contradicting, then all new max-blocks inserted between $\vec{A}_1$ and $\vec{A}_2$ must be cn-equal to them both.

  Assume that $\vec{A}_1$ and $\vec{A}_2$ are not both max-blocks in $\vec{O}_M$. Without loss of generality, let $\vec{A}_1$ be member of a larger max-block in $\vec{O}_M$. We show that $\vec{A}_2$ must also participate in the same max-block.

  Since $\vec{A}_1$ joined a larger block, there must exist another block, $\vec{B}$, adjacent to $\vec{A}_1$, such that their pair is cn-inverted. Let $\vec{B}$ stand to the left of $\vec{A}_1$ (in the opposite case, the proof is similar): $\vec{O}_M = \vec{X}\|\vec{B}\|\vec{A}_1\|\vec{Y}\|\vec{A}_2\|\vec{Z}$. The pair $\langle\vec{B},\vec{A}_1\rangle$ is cn-inverted, i.e., $cn(\vec{B})|_{\mathcal{B}(M)\cup\vec{X}} > cn(\vec{A}_1)|_{\mathcal{B}(M)\cup\vec{X}\cup\vec{B}}$. From Lemma 9, $cn(\vec{B}\|\vec{A}_1)|_{\mathcal{B}(M)\cup\vec{X}} > cn(\vec{A}_1)|_{\mathcal{B}(M)\cup\vec{X}\cup\vec{B}}$, and we must add to the block $\vec{B}\|\vec{A}_1$ all blocks from $\vec{Y}$, because they are all cn-equal to $\vec{A}_1$. Also, $cn(\vec{A}_1)|_{\mathcal{B}(M)\cup\vec{X}\cup\vec{B}} = cn(\vec{A}_2)|_{\mathcal{B}(M)\cup\vec{X}\cup\vec{B}\cup\vec{A}_1}$, and $\vec{A}_2$ must also be added to the block. Thus, $\vec{A}_1$ and $\vec{A}_2$ belong to the same max-block in $\vec{O}_M$.

- If $\vec{A}_1$ and $\vec{A}_2$ belong to the same max-block in $\vec{O}'_M$, then this block is either violated in $\vec{O}_M$, or not. In the former case, $\vec{O}_M$ is MC-contradicting, by Corollary 3. In the latter case, $\vec{A}_1$ and $\vec{A}_2$ belong to the same max-block in $\vec{O}_M$.

- If $\vec{O}'_M$ is MC-contradicting, then $\vec{O}_M$ is MC-contradicting too (the proof is easy). □

Now we can prove Lemma 8:

**Lemma 8**
Let $\mathcal{S}_0$ be a set of subgoals, $N$ be a node in the divisibility tree of $\mathcal{S}_0$ and $\vec{O}_N = \vec{Q}\|\vec{A}_1\|\vec{A}_2\|\vec{R}$ be an ordering of $\mathcal{S}(N)$, where $\vec{A}_1$ and $\vec{A}_2$ are max-blocks, mutually independent and cn-equal under the bindings of $\mathcal{B}(N) \cup \vec{Q}$. Then $\vec{O}_N$ is blockwise-equivalent with $\vec{O}'_N = \vec{Q}\|\vec{A}_2\|\vec{A}_1\|\vec{R}$.

**Proof:**

Let $\vec{S}$ be a minimal ordering of $\mathcal{S}_0$ binder-consistent with $\vec{O}_N$. By Corollary 3, $\vec{S}$ does not violate the blocks of $\vec{O}_N$, in particular $\vec{A}_1$ and $\vec{A}_2$: $\vec{S} = \vec{X}\|\vec{A}_1\|\vec{Y}\|\vec{A}_2\|\vec{Z}$. Let $\vec{S}' = \vec{S}|_{\vec{O}'_N}^{\vec{O}'_N} = \vec{X}\|\vec{A}_2\|\vec{Y}\|\vec{A}_1\|\vec{Z}$. We must show that $\vec{S}'$ is minimal, which implies blockwise equivalence of $\vec{O}_N$ and $\vec{O}'_N$.

If $\vec{Y}$ is empty, then $Cost(\vec{S}) = Cost(\vec{S}')$ by Lemma 2 ($\vec{A}_1$ and $\vec{A}_2$ are adjacent, mutually independent and cn-equal; thus, their transposition does not change the cost).

If $\vec{Y}$ is not empty, then by Corollary 2 $\vec{Y}$ is mutually independent of both $\vec{A}_1$ and $\vec{A}_2$ ($\vec{S}$ is binder-consistent with $\vec{O}_N$, therefore $\mathcal{B}(N) \subseteq \vec{X}$, and consequently $\vec{Y} \cap \mathcal{B}(N) = \emptyset$). $\vec{Y}$ can be divided into several blocks, each one of them cn-equal to $\vec{A}_1$ and $\vec{A}_2$: since $\vec{S}$ is minimal, $\vec{O}_N$ cannot be MC-contradicting, and the claim follows from Lemma 10. By Lemma 9, $cn(\vec{Y})|_{\vec{X}} = cn(\vec{A}_1)|_{\vec{X}} = cn(\vec{A}_2)|_{\vec{X}}$. By Lemma 2:

$$
\begin{aligned}
Cost(\vec{S}) &= Cost(\vec{X}\|\vec{A}_1\|\vec{Y}\|\vec{A}_2\|\vec{Z}) &=& \quad // \; swap(Y, A_2) \\
&= Cost(\vec{X}\|\vec{A}_1\|\vec{A}_2\|\vec{Y}\|\vec{Z}) &=& \quad // \; swap(A_1, A_2) \\
&= Cost(\vec{X}\|\vec{A}_2\|\vec{A}_1\|\vec{Y}\|\vec{Z}) &=& \quad // \; swap(A_1, Y) \\
&= Cost(\vec{X}\|\vec{A}_2\|\vec{Y}\|\vec{A}_1\|\vec{Z}) &=& \quad Cost(\vec{S}')
\end{aligned}
$$

Minimality of $\vec{S}'$ implies blockwise equivalence of $\vec{O}_N$ and $\vec{O}'_N$. □

## Appendix B. Correctness of the DAC Algorithm

In this section we show that the DAC algorithm is correct, i.e., given a set of subgoals $\mathcal{S}_0$, it returns its minimal ordering. It suffices to show that the candidate set of the root node of $DTree(\mathcal{S}_0, \emptyset)$ is valid. In such a case, as follows from the definition of valid sets, it must contain a minimal ordering. The algorithm returns one of the cheapest candidates of the root. Therefore, if the candidate set of the root is valid, the DAC algorithm must return a minimal ordering of $\mathcal{S}_0$.

We start by defining *strong validity* of sets of orderings. We then prove that strong validity implies validity. Finally, we use induction to prove a theorem, showing that the candidate set produced for each node in the divisibility tree is strongly valid.

**Definition:** Let $\mathcal{S}_0$ be a set of subgoals, $N$ be a node in the divisibility tree of $\mathcal{S}_0$. The set $\mathcal{C}_N \subseteq \pi(\mathcal{S}(N))$ is *strongly valid*, if every ordering in $\pi(\mathcal{S}(N)) \setminus \mathcal{C}_N$ is either MC-contradicting or blockwise-equivalent to some member of $\mathcal{C}_N$, unless no ordering of $\mathcal{S}(N)$ is min-consistent.

$StronglyValid_{N,\mathcal{S}_0}(\mathcal{C}_N) \iff$
$[\exists \vec{O}'_N \in \pi(\mathcal{S}(N)): \; MC_{N,\mathcal{S}_0}(\vec{O}'_N)] \rightarrow [\vec{O}_N \in \pi(\mathcal{S}(N)) \setminus \mathcal{C}_N \rightarrow MCC_{N,\mathcal{S}_0}(\vec{O}_N) \vee$
$(\exists \vec{O}''_N \in \mathcal{C}_N \wedge MCE_{N,\mathcal{S}_0}(\vec{O}_N, \vec{O}''_N))]$

**Lemma 11** *A strongly valid set of orderings is valid.*

**Proof:** Let $\mathcal{S}_0$ be a set of subgoals, $N$ be a node in the divisibility tree of $\mathcal{S}_0$, $\mathcal{C}(N)$ be a strongly valid set of orderings of $N$.

If there is no min-consistent ordering of $N$, then $\mathcal{C}(N)$ is valid, by the definition of a valid set (Section 4.2).

Otherwise, there exists at least one minimal ordering of $\mathcal{S}_0$, binder-consistent with $N$. Every ordering in $\pi(\mathcal{S}(N)) \setminus \mathcal{C}(N)$ is either MC-contradicting or blockwise-equivalent to some member of $\mathcal{C}(N)$. To prove that $\mathcal{C}(N)$ is valid, we must show that it contains an ordering $\vec{O}_N$, which is binder-consistent with some minimal ordering $\vec{S}$ of $\mathcal{S}_0$.

Let $\vec{S}'$ be a minimal ordering of $\mathcal{S}_0$, binder-consistent with $N$. Let $\vec{O}'_N$ be the projection of $\vec{S}'$ on $N$. If $\vec{O}'_N \in \mathcal{C}(N)$, we are done ($\vec{O}_N = \vec{O}'_N$, $\vec{S} = \vec{S}'$). Otherwise, $\vec{O}'_N \in \pi(\mathcal{S}(N)) \setminus \mathcal{C}(N)$. $\vec{O}'_N$ cannot be MC-contradicting (it is min-consistent to $\vec{S}'$), therefore it must be blockwise-equivalent to some $\vec{O}''_N \in \mathcal{C}(N)$. Blocks of $\vec{O}'_N$ are not violated in $\vec{S}'$, since $\vec{S}'$ is minimal (Corollary 3). Therefore the substitution $\vec{S}'' = \vec{S}'|_{\vec{O}'_N}^{\vec{O}''_N}$ is well defined. $\vec{S}''$ is minimal, since $\vec{S}'$ is minimal and $\vec{O}'_N$ and $\vec{O}''_N$ are blockwise-equivalent. $\vec{S}''$ is binder-consistent with $\vec{O}''_N$, since $\vec{S}'$ was binder-consistent with $\vec{O}'_N$. Thereupon $\vec{S}''$ and $\vec{O}''_N$ satisfy the requirements of validity ($\vec{O}_N = \vec{O}''_N$, $\vec{S} = \vec{S}''$). $\qquad\square$

### Theorem 3
*Let $\mathcal{S}_0$ be a set of subgoals. For each node $N$ of the divisibility tree of $\mathcal{S}_0$, Algorithm 6 creates a strongly valid candidate set of orderings.*

**Proof:** By induction on the height of $N$'s subtree.

**Inductive base:** $N$ is a leaf node, which means that $\mathcal{S}(N)$ is independent under $\mathcal{B}(N)$. The candidate set of $N$ contains one element, whose subgoals are sorted by $cn$. All orderings that belong to $\pi(\mathcal{S}(N)) \setminus CandSet(N)$ are either not sorted by $cn$, and hence are MC-contradicting (Lemma 4), or are sorted by $cn$, and hence are blockwise-equivalent to the candidate (Corollary 4). Consequently, $CandSet(N)$ is strongly valid.

**Inductive hypothesis:** For all children of $N$, Algorithm 6 produces strongly valid candidate sets.

**Inductive step:** An internal node in a divisibility tree is either an AND-node or an OR-node.

1. $N$ is an AND-node. Let $N_1, N_2, \dots N_k$ be the children of $N$. First we show that $ConsSet(N)$ is strongly valid.

   Let $\vec{O}_N \in \pi(\mathcal{S}(N)) \setminus ConsSet(N)$. For all $1 \le i \le k$, let $\vec{O}_i$ be the projection of $\vec{O}_N$ on $N_i$. The set of projections $\{\vec{O}_1, \vec{O}_2, \dots \vec{O}_k\}$ can belong to one of the three following types, with regard to $\vec{O}_N$.

   (a) The sets of the first type contain at least one MC-contradicting projection. In such a case $\vec{O}_N$ is MC-contradicting too. Assume the contrary: there exists a minimal ordering $\vec{S}$ of $\mathcal{S}_0$, binder-consistent with $\vec{O}_N$. Let $\vec{O}_i$ be an MC-contradicting projection. Since $\vec{O}_i$ is consistent with $\vec{O}_N$, it is also consistent with $\vec{S}$. Since $\mathcal{B}(N_i) = \mathcal{B}(N)$, all subgoals of $\mathcal{B}(N_i)$ appear in $\vec{S}$ before subgoals of $\mathcal{S}(N_i)$. Therefore, $\vec{O}_i$ is binder-consistent with $\vec{S}$, and since $\vec{S}$ is minimal, $\vec{O}_i$ is min-consistent and not MC-contradicting – a contradiction.

(b) The sets of the second type do not contain MC-contradicting projections, but in $\vec{O}_N$ some block of some projection is violated, or max-blocks from different projections are not ordered by $cn$. In such a case, $\vec{O}_N$ is MC-contradicting, by Corollary 3 and Lemma 6.

(c) The sets of the third type do not contain MC-contradicting projections, and max-blocks of the projections are not violated in $\vec{O}_N$ and are sorted by $cn$. Every projection $\vec{O}_i$ either belongs to $CandSet(N_i)$, or not. If $\vec{O}_i \notin CandSet(N_i)$, then there exists $\vec{O}'_i \in CandSet(N_i)$ such that $\vec{O}_i$ is blockwise-equivalent to $\vec{O}'_i$ (because $CandSet(N_i)$ is strongly valid by the inductive hypothesis, and $\vec{O}_i$ is not MC-contradicting). If $\vec{O}_i \in CandSet(N_i)$, we can set $\vec{O}'_i = \vec{O}_i$.

Let $\vec{O}'_N = \vec{O}_N|_{\vec{O}_1}^{\vec{O}'_1}|_{\vec{O}_2}^{\vec{O}'_2} \dots |_{\vec{O}_k}^{\vec{O}'_k}$. This substitution is well defined, since each $\vec{O}_i$ has the same number of max-blocks as $\vec{O}'_i$, and max-blocks of the projections are not violated in $\vec{O}_N$. Let $\vec{S}$ be a minimal ordering of $\mathcal{S}_0$, binder-consistent with $\vec{O}_N$. Since $\vec{S}$ is minimal, blocks of $\vec{O}_1$ are not violated in $\vec{S}$. Since $\vec{O}_1$ is blockwise-equivalent to $\vec{O}'_1$, the ordering $\vec{S}_1 = \vec{S}|_{\vec{O}_1}^{\vec{O}'_1}$ is well-defined and minimal. In $\vec{S}_1$ the positions of the subgoals from $\mathcal{B}(N)$ did not change; thus, $\vec{O}_2$ is min-consistent with $\vec{S}_1$, and blockwise equivalence of $\vec{O}_2$ and $\vec{O}'_2$ entails minimality of the ordering $\vec{S}_2 = \vec{S}_1|_{\vec{O}_2}^{\vec{O}'_2} = \vec{S}|_{\vec{O}_1}^{\vec{O}'_1}|_{\vec{O}_2}^{\vec{O}'_2}$. We continue with other $\vec{O}_i$-s, and finally obtain that $\vec{S}' = \vec{S}|_{\vec{O}_1}^{\vec{O}'_1}|_{\vec{O}_2}^{\vec{O}'_2} \dots |_{\vec{O}_k}^{\vec{O}'_k}$ is minimal. From the definition of $\vec{O}'_N$, $\vec{S}' = \vec{S}|_{\vec{O}_N}^{\vec{O}'_N}$ (note that we introduced blockwise equivalence and strong validity only to be able to perform this transition). $\vec{S}'$ is minimal, therefore $\vec{O}_N$ is blockwise-equivalent to $\vec{O}'_N$. $\vec{O}'_N \in ConsSet(N)$, since all its projections are candidates of the child nodes. Thereupon, $\vec{O}_N$ is blockwise-equivalent to a member of $ConsSet(N)$.

So, $ConsSet(N)$ is strongly valid. To prove that $CandSet(N)$ is strongly valid, it suffices to show that all the members of $ConsSet(N)$ that are not included in $CandSet(N)$ by Algorithm 6, are either MC-contradicting or blockwise-equivalent to members of $CandSet(N)$. Such orderings can be of three types:

(a) Orderings that violate blocks of the children projections. They are MC-contradicting by Corollary 3.

(b) Orderings that do not violate blocks, but where max-blocks of children projections are not ordered by $cn$. They are MC-contradicting by Lemma 6.

(c) Orderings that do not violate blocks and have them sorted by $cn$. For each combination of projections, one consistent ordering of $N$ is retained in the candidate set, and all the other are rejected. By Corollary 5, the rejected orderings are blockwise-equivalent to the retained candidate.

Consequently, $CandSet(N)$ is strongly valid.

2. $N$ is an OR-node. Again, we start with showing that $ConsSet(N)$ is strongly valid.

Let $\vec{O}_N \in \pi(\mathcal{S}(N)) \setminus ConsSet(N)$. $\vec{O}_N$ is constructed from a binder $H$ and a "tail" sequence $\vec{T}$: $\vec{O}_N = H \| \vec{T}$. Let $N_H$ be the child of $N$ that corresponds to the binder $H$. By the inductive hypothesis, $CandSet(N_H)$ is strongly valid. $\vec{T} \notin CandSet(N_H)$, since otherwise $\vec{O}_N \in ConsSet(N)$. Therefore, $\vec{T}$ is either MC-contradicting, or blockwise-equivalent to some $\vec{T}' \in CandSet(N_H)$. If $\vec{T}$ is MC-contradicting, $\vec{O}_N$ is MC-contradicting too (proof by contradiction, as for AND-nodes). If $\vec{T}$ is blockwise-equivalent to $\vec{T}'$, then $\vec{O}_N = H \| \vec{T}$ is blockwise-equivalent to $H \| \vec{T}' \in ConsSet(N)$ (the proof is easy). Hence, $ConsSet(N)$ is strongly valid. The only orderings of $ConsSet(N)$ that are not included in $CandSet(N)$ by the DAC algorithm have cheaper permutations of their leading max-blocks, and therefore are MC-contradicting, by Lemma 7. Hence, $CandSet(N)$ is strongly valid. □

**Corollary 7** *The candidate set found by Algorithm 6 for the root node is valid.*

**Corollary 8** *Algorithm 6 finds a minimal ordering of the given set of subgoals.*